\def\year{2020}\relax
\documentclass[letterpaper]{article} 
\usepackage{aaai20}  
\usepackage{times}  
\usepackage{helvet} 
\usepackage{courier}  
\usepackage[hyphens]{url}  
\usepackage{graphicx} 
\usepackage{graphicx}  
\usepackage{amsmath,amssymb,amsfonts}
\usepackage{booktabs}
\usepackage{natbib}
\usepackage{algorithm}
\usepackage[noend]{algpseudocode}
\usepackage{graphicx}
\usepackage{tabularx}
\usepackage{textcomp}
\usepackage{xcolor}
\usepackage{color}
\usepackage{subcaption}
\usepackage[draft]{hyperref}

\begin{document}

\title{Regional Tree Regularization for Interpretability in Deep Neural Networks}
\author{
    Mike Wu\textsuperscript{\rm 1},
    Sonali Parbhoo\textsuperscript{\rm 2,3},
    Michael C. Hughes\textsuperscript{\rm 4}, \\
    {\bf \Large
    Ryan Kindle,
    Leo Celi\textsuperscript{\rm 6},
    Maurizio Zazzi\textsuperscript{\rm 8},
    Volker Roth\textsuperscript{\rm 2},
    Finale Doshi-Velez\textsuperscript{\rm 3}} \\
    \textsuperscript{\rm 1} Stanford University, wumike@stanford.edu \\
    \textsuperscript{\rm 2} University of Basel,
    {volker.roth}@unibas.ch \\
    \textsuperscript{\rm 3} Harvard University SEAS, \{sparbhoo, finale\}@seas.harvard.edu \\
    \textsuperscript{\rm 4} Tufts University, michael.hughes@tufts.edu \\
    \textsuperscript{\rm 6} Massachusetts Institute of Technology, lceli@mit.edu \\
    \textsuperscript{\rm 8} University of Siena,
    maurizio.zazzi@unisi.it \\
}
\maketitle

\begin{abstract}
The lack of interpretability remains a barrier to adopting deep neural networks across many safety-critical domains. Tree regularization was recently proposed to encourage a deep neural network's decisions to resemble those of a globally compact, axis-aligned decision tree. However, it is often unreasonable to expect a single tree to predict well across all possible inputs. In practice, doing so could lead to neither interpretable nor performant optima.
To address this issue, we propose \emph{regional} tree regularization -- a method that encourages a deep model to be well-approximated by several separate decision trees
specific to predefined regions of the input space.
Across many datasets, including two healthcare applications, we show our approach delivers simpler explanations than other regularization schemes without compromising accuracy. Specifically, our regional regularizer finds many more ``desirable" optima compared to global analogues.
\end{abstract}

\section{Introduction}
Deep neural networks have become state-of-the-art in many applications, and are poised to advance prediction in real-world domains such as healthcare \citep{miotto2016deep, gulshan2016development}.
However, understanding when a model's outputs can be trusted and how the model might be improved remains a challenge in safety-critical domains. \citet{chen2017machine} discuss how these challenges inhibit the adoption of deep models in clinical healthcare. Without interpretability, humans are unable to incorporate domain knowledge and effectively audit predictions.

Prior work for explaining deep models has focused on two types of explanation: \emph{global} and \emph{local}. A global explanation (e.g. \citet{wu2018beyond, che2015distilling}) returns a single explanation for the \emph{entire} model. However, if the explanation is simple enough to understand, it is unlikely to be faithful to the model across all inputs.  In contrast, local explanation (e.g. \citet{ribeiro2016should, selvaraju2016grad}) explain predictions for an isolated input, which may miss larger patterns. Local approaches also leave ambiguous whether the logic for an input $x$ applies to a nearby input $x'$, which can lead to poor assumptions about generalizability.

Our work marks a major departure from this previous literature.
We introduce a middle-ground between global and local approaches: \textit{regional} explanations.
Given a predefined set of human-intuitive regions of input space, we require the explanation for \emph{each} region to be simple.  This idea of regions is consistent with \emph{context-dependent} human reasoning~\citep{miller2018explanation}. For example, physicians in the intensive care unit do not expect treatment rules to be the same across patients of different risk levels.  By requiring \emph{all} regional explanations to be simple, we prevent the model from being simple in one region to make up for complexity in another (something global explanation methods cannot do).



Our operational definition of interpretability is to make the explanation for each region easily \emph{human-simulable}. Simulable explanations allows humans to, ``in reasonable time, combine inputs and explanation to produce outputs, forming a foundation for auditing and correcting predictions" \citep{lipton2016mythos}.
Like~\citet{wu2018beyond}, we select decision trees as a simulable surrogate prediction model and develop a joint optimization objective that balances prediction error with a penalty on the size of a  deep model's surrogate tree.
However, our training objective requires a separate tree for each region, rather than the entire input space.
Decomposing into regions provides for a more flexible deep model while still revealing  prediction logic that can be understood by humans. However, inference for regionally simulable explanations is more challenging than the global case.

Our technical contributions are twofold. First, we introduce a new regularization penalty term that ensures simplicity across \emph{all} regions
while being tractable for gradient-based optimization.
Second, we develop concrete innovations to improve the stability of optimization, which are essential to making our new regularization term effective in practice. These last innovations would lead to improvements to global tree regularization (e.g. \citet{wu2018beyond}) as well.
We achieve comparable performance to complex models while learning a much simpler decision function.
Through exposition and careful experiments, we emphasize that our regional penalization is \emph{distinct from} and \emph{better than} simply using a global tree regularization constraint~\citep{wu2018beyond} where the root of the tree divides examples by region.


\section{Related Work}

\paragraph{Global Interpretability}
Many approaches exist to summarize a \emph{trained} black box model.  Works such as \citep{mordvintsev2015inceptionism} expose the features a representation encodes but not the logic. \citet{amir2018highlights} and \citet{kim2014bayesian} provide an informative set of examples that represent what the system has learned. Recently, Activation maximisation of neural networks \citep{montavon2018methods} tries to find input patterns that produce the maximum response for a quantity of interest. Similarly, model distillation compresses a source network into a smaller target neural network \citep{frosst2017distilling}.
Likewise, Layerwise-Relevance Propagation (LRP) \citep{binder2016layer,bach2015pixel} produces a heatmap of relevant information for prediction based on aggregating the weights of a neural network. \citet{guan2019towards} improve on LRP with a similar measure for NLP models that better capture coherency and generality.
However, these summaries have no guarantees of being simulable since an expert cannot necessarily step through any calculation that produces a decision.

\paragraph{Local Interpretability}
In contrast, local approaches provide explanation for a specific input. \citet{ribeiro2016should} show that using the weights of a sparse linear model, one can explain the decisions of a black box model in a small area near a fixed data point.
Similarly, \citet{singh2016programs} and \citet{koh2017understanding} output a simple program or an influence function, respectively. Other approaches have used input gradients (which can be thought of as infinitesimal perturbations) to characterize  local logic \citep{maaten2008visualizing,selvaraju2016grad}. However, such local explanations do not match with human notions of contexts \citep{miller2018explanation}: a user may have difficulty knowing if and when explanations generated locally for input $x$ translate to a new input $x'$.

\paragraph{Optimizing for Interpretability}
Few works include interpretability as an optimization objective \emph{during} model training, rather than attempt explanation on an already trained model.  \citet{ross2017right, wu2018beyond} include regularizers that capture explanation properties (which are input gradients and decision trees, respectively). \citet{krening2017learning} jointly train an image classifier alongside a captioning model to provide a verbal explanation for any prediction; although not simulable, the generated text influences the weights for the image network during training.  In this work, we optimize for ``regional" simulability, which we show to find more interpretable optima than optimizing for many measures of global simulability.

\section{Background and Notation}
We consider supervised learning tasks given a dataset of $N$ labeled examples, $\mathcal{D} = \{(\mathbf{x}_n, \mathbf{y}_n)\}_{n=1}^N$, with continuous inputs $\mathbf{x} \in \mathcal{X}^P$ and binary outputs $\mathbf{y} \in \{0,1\}^Q$. Define a predictor $\hat{\mathbf{y}}_n = f(\mathbf{x}_n; \theta)$ as a multilayer perceptron (MLP), denoted by $f(\cdot; \theta)$. The parameters $\theta$ are trained to minimize
\begin{equation}
\arg\min_{\theta\in\Theta} \sum_{n=1}^{N} \mathcal{L}(\mathbf{y}_n, f(\mathbf{x}_n; \theta)) + \lambda \Omega(\theta)
\label{eqn:objective}
\end{equation}
where $\Omega(\theta)$ represents a regularization penalty with scalar strength $\lambda \in \mathbb{R}^{+}$.  Common regularizers include the L$_1$ or L$_2$ norm of $\theta$, or our new regional regularizer. We shall refer to $f(\cdot; \theta)$ as a \textit{target neural model}.

\paragraph{Global Tree Regularization}
\citet{wu2018beyond} introduce a regularizer that penalizes models for being hard to simulate, where simulability is the complexity of a single, global decision tree that approximates the target neural model. They define tree complexity as the \textit{average decision path length} (APL), or the expected number of binary decisions (each one corresponding to a node within the tree) that must be stepped through to produce a prediction. We compute the APL of a predictor $f$ given a dataset of size $N$ as:
\begin{equation}
 \Omega^{\texttt{global}}(\theta) \triangleq
 \textsc{APL}(\{\mathbf{x}_n\}_{n=1}^N, f(\cdot, \theta))
 \label{eqn:reg:naive}
\end{equation}
The $\textsc{APL}$ procedure is defined in Alg.~\ref{algorithm:2}, where the subroutine $\textsc{TrainTree}$ fits a decision tree (e.g. CART).
The $\textsc{GetDepth}$ subroutine returns the depth of the leaf node predicted by the tree given an input example $\mathbf{x}_n$.

\begin{algorithm}[!t]
\caption{\textsc{APL}~\citep{wu2018beyond}}
\begin{algorithmic}[1]
\Require{
  \Statex $f(\cdot, \theta)$: prediction function, with parameters $\theta$
  \Statex $\{ \mathbf{x}_i \}_{i=1}^{N}$: a set of $N$ input examples
}
\Function{$\textsc{APL}$}{$\{ \mathbf{x}_i \}_{i=1}^{N}, f, h$}
  \State $\hat{\mathbf{y}}_i = f(\mathbf{x}_{i}, \theta)$, $\forall i \in \{1, 2, \ldots N\}$
  \State $T = \textsc{TrainTree}( \{ \mathbf{x}_{i}, \hat{\mathbf{y}}_i \}_{i=1}^N)$
  \State \Return $\text{mean}( \{ \textsc{GetDepth}(T, \mathbf{x}_i) \}_{i=1}^N )$
\EndFunction
\end{algorithmic}
\label{algorithm:2}
\end{algorithm}

Importantly, $\textsc{TrainTree}$ is not differentiable, making optimization of Eq.~\eqref{eqn:reg:naive} challenging using gradient descent methods. To overcome this challenge, \cite{wu2018beyond} introduce a surrogate regularizer $\hat\Omega^{\texttt{global}}(\theta)$, which is a differentiable function that \emph{estimates} the target neural model's APL for a specific parameter vector $\theta$. In practice, $\hat{\Omega}^{\texttt{global}}(\theta)$ is a small multi-layer perceptron with weight and bias parameters $\phi$, which we refer to as the \textit{surrogate model}.

Training the surrogate model to produce accurate APL estimates is a supervised learning problem.
First, collect a dataset of $J$ parameter values and associated APL values:  $\mathcal{D}^\theta = \{\theta_j, \Omega^{\texttt{global}}(\theta_j)\}_{j=1}^J$.
Parameter examples, $\theta_j$, can be gathered from every update step of training the target neural model. Next, train the surrogate MLP by minimizing the sum of squared errors:
\begin{equation}
\arg\min_{\phi} \textstyle \sum_{j=1}^J (\Omega^{\texttt{global}}(\theta_j) - \hat{\Omega}^{\texttt{global}}(\theta_j; \phi))^2
\label{eqn:surrogate}
\end{equation}
Optimizing even a single surrogate can be challenging; for our case, we need to train and maintain \emph{multiple} surrogates.


\section{Regionally Faithful Explanations}
\label{sec:treereg}
Global summaries such as \cite{wu2018beyond} face a tough trade-off between simulability and accuracy.
A strong penalty on simulability will force the target network to be too simple and lose accuracy. However, ignoring simulability may produce an accurate target network that is incomprehensible to humans (note that the optimization procedure forces the decision tree explanation to be faithful to the network).  But do we need a single explanation for the whole model?  The cognitive science literature tells us that people build context-dependent models of the world; they do not expect the same rule to apply in all circumstances \citep{miller2018explanation}. For instance, doctors may use different models for treating patients depending on their predicted level of risk or whether the subject has just come out of surgery or not.

Building on this notion, we consider the problem in which we are given a collection of $R$ regions that cover the entire input space: $\mathcal{X}_1, \ldots \mathcal{X}_R$, where $\cup_{r=1}^R \mathcal{X}_r \subseteq \mathcal{X}^P$. We do not require these regions to be disjoint nor tile the full space. We emphasize that it is \emph{essential} that these regions correspond to human-understandable categories (e.g. surgery vs. non-surgery patients)
to avoid confusion about when each explanation applies.  For now, we assume that the regions are fully specified in advance, likely by domain experts.
We could also use interactive, interpretable clustering methods (e.g. \citet{chuang2014human,kim2015interactive}) to group the input space in a data-driven way.

\begin{figure}[t]
  \centering
  \begin{subfigure}[b]{0.24\linewidth}
    \includegraphics[width=\linewidth]{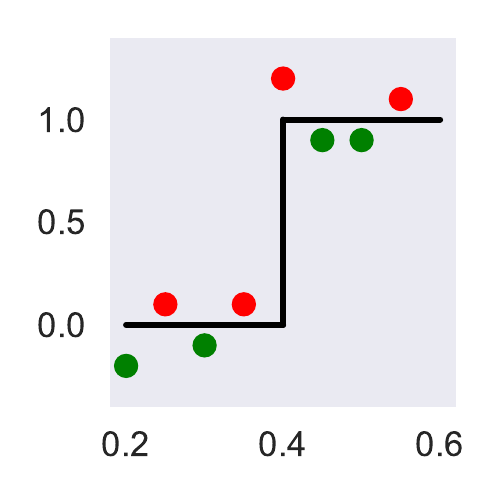}
    \caption{True}
  \end{subfigure}
  \begin{subfigure}[b]{0.24\linewidth}
    \includegraphics[width=\linewidth]{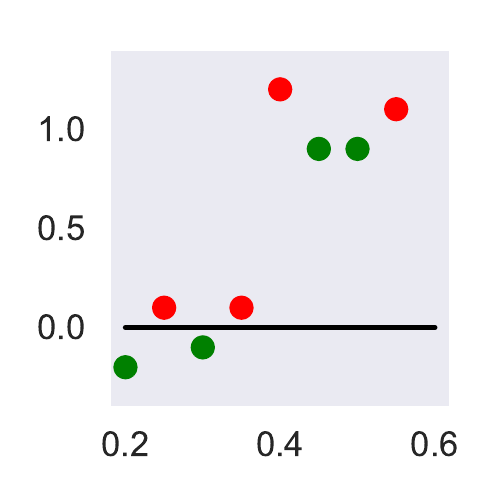}
    \caption{Global}
  \end{subfigure}
  \begin{subfigure}[b]{0.24\linewidth}
    \includegraphics[width=\linewidth]{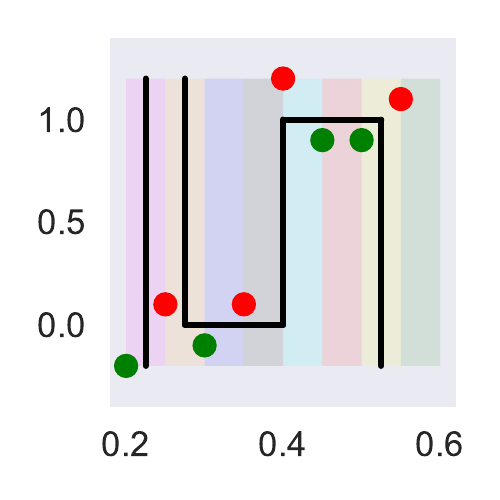}
    \caption{Local}
  \end{subfigure}
  \begin{subfigure}[b]{0.24\linewidth}
    \includegraphics[width=\linewidth]{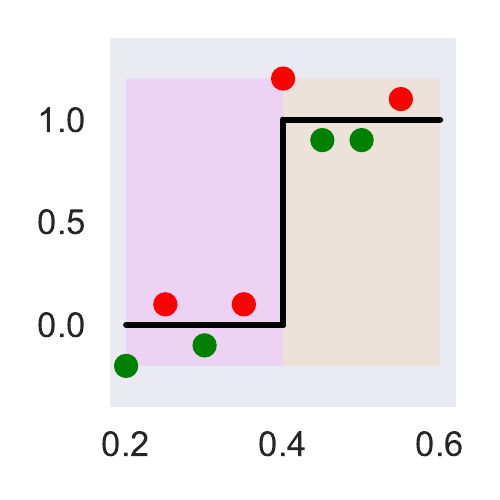}
    \caption{Regional}
  \end{subfigure}
  \caption{Decision boundaries learned by global (b), local (c), and
    regional (d) tree regularization. (a) shows the true decision boundary. Red and green points
    represent the training dataset. Lightly colored areas represent
    regions.
    }
  \label{fig:toy2}
  \vspace{-1em}
\end{figure}

Given the regions, our goal is to find a high-performing target network that is simple in \emph{every} region.
Fig.~\ref{fig:toy2} highlights the distinctions between global, local, and regional tree regularization on a 2D toy dataset with binary labels. The true decision boundary abruptly changes at one specific value of the first input dimension (shown on the x-axis). We intend there to be two ``regions'', representing input values below and above this threshold.
The key question is what inductive biases do different regularization strategies impose.
Global regularization (b) commonly imposes ``simplicity" at the cost of accuracy under strong regularization. Local regularization (c) produces simple boundaries around each data point but a complex global boundary.
Our regional regularization (d) over two regions recovers the expected boundary.

\subsection{L1 Regional Tree Regularization: A Failed Attempt}
A naive way to generalize global tree regularization to regions is to penalize the sum of the APLs in each region:
\begin{equation}
  \Omega^{\texttt{regional-L}1}(\theta)
  \triangleq \sum_{r=1}^R \textsc{APL}(X_r, f(\cdot, \theta))
\label{eqn:naive-apl}
\end{equation}
where $\textsc{APL}(\cdot)$ is as defined in Alg.~\ref{algorithm:2}, $f$ is the target neural model,
and $X_r = \{x_n : x_n \in \mathcal{X}_r\}$ denotes training data in region $r$.
If the regions form a set partition of $\mathcal{X}^P$ and all regions contain equal amounts of input data, this regularizer is essentially equivalent to global tree regularization~\citep{wu2018beyond} where the root decision node is constrained to split by region. We refer to this as L$_1$ \textit{regional tree regularization}.

The trouble with this naive solution is that smaller or simpler regions may be over-regularized (possibly to trivial functions) in order to minimize the sum while other regions stay complex. Fig.~\ref{fig:toyexp3}(a,b,c) shows this effect: the two regions have different complexities of boundaries, but the naive metric (b,c) over-simplifies to minimize Eq.~\eqref{eqn:naive-apl}.


\subsection{L0 Regional Tree Regularization: Our Proposal}
\begin{figure}[t]
  \centering
  \begin{subfigure}[b]{0.19\linewidth}
    \includegraphics[width=\linewidth]{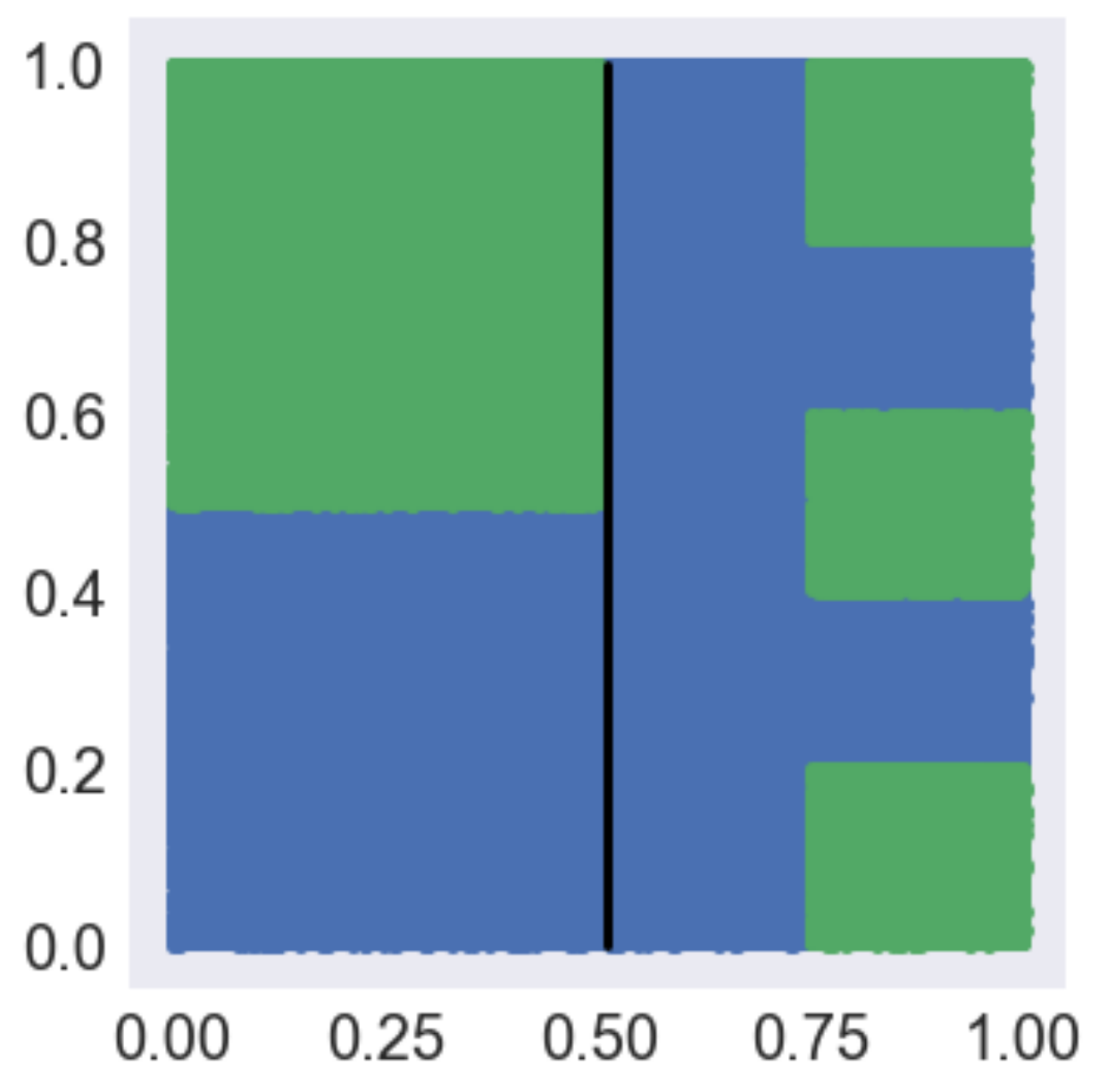}
    \caption{True}
  \end{subfigure}
  \begin{subfigure}[b]{0.19\linewidth}
    \includegraphics[width=\linewidth]{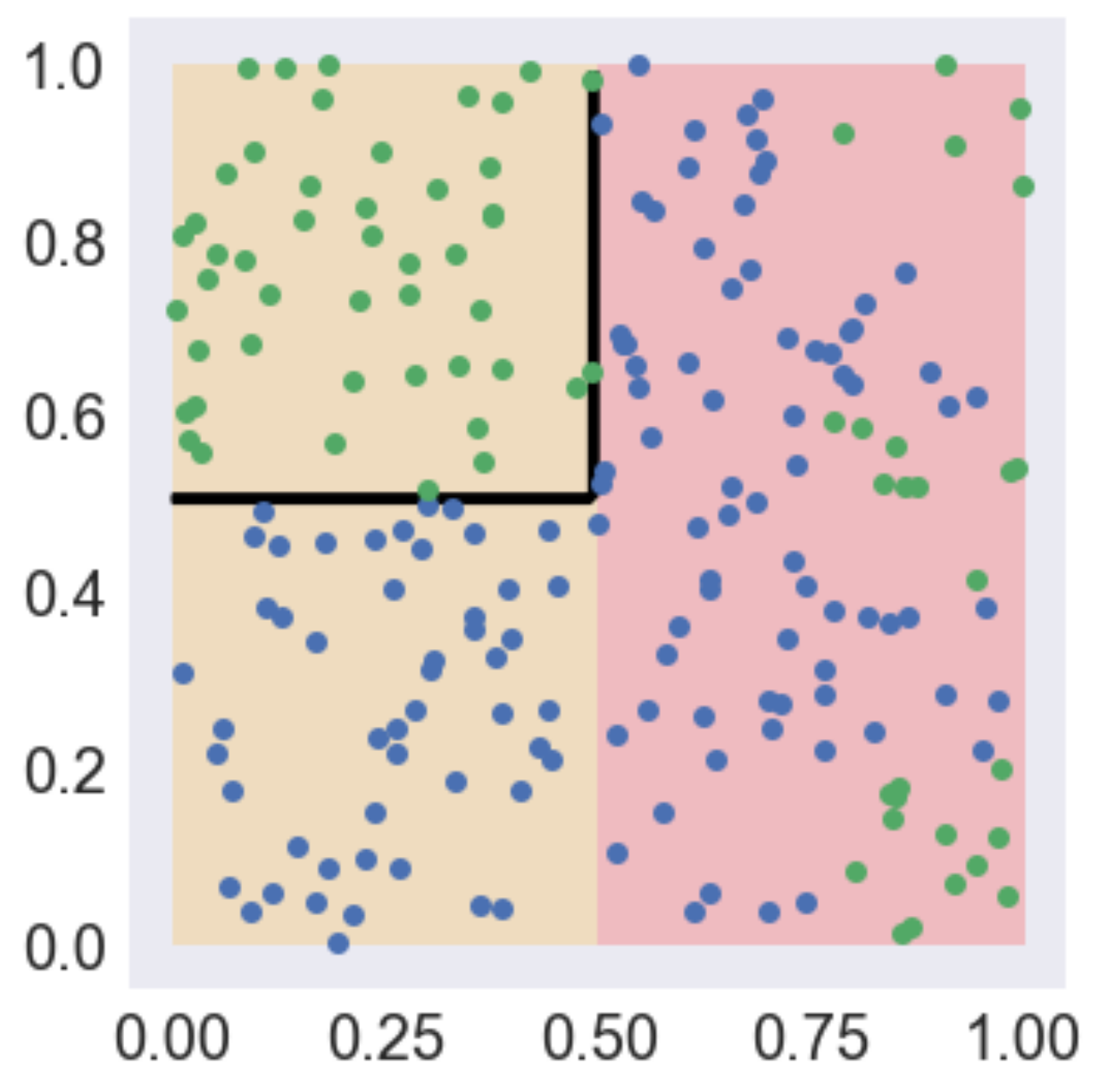}
    \caption{L$_1$ run A}
  \end{subfigure}
  \begin{subfigure}[b]{0.19\linewidth}
    \includegraphics[width=\linewidth]{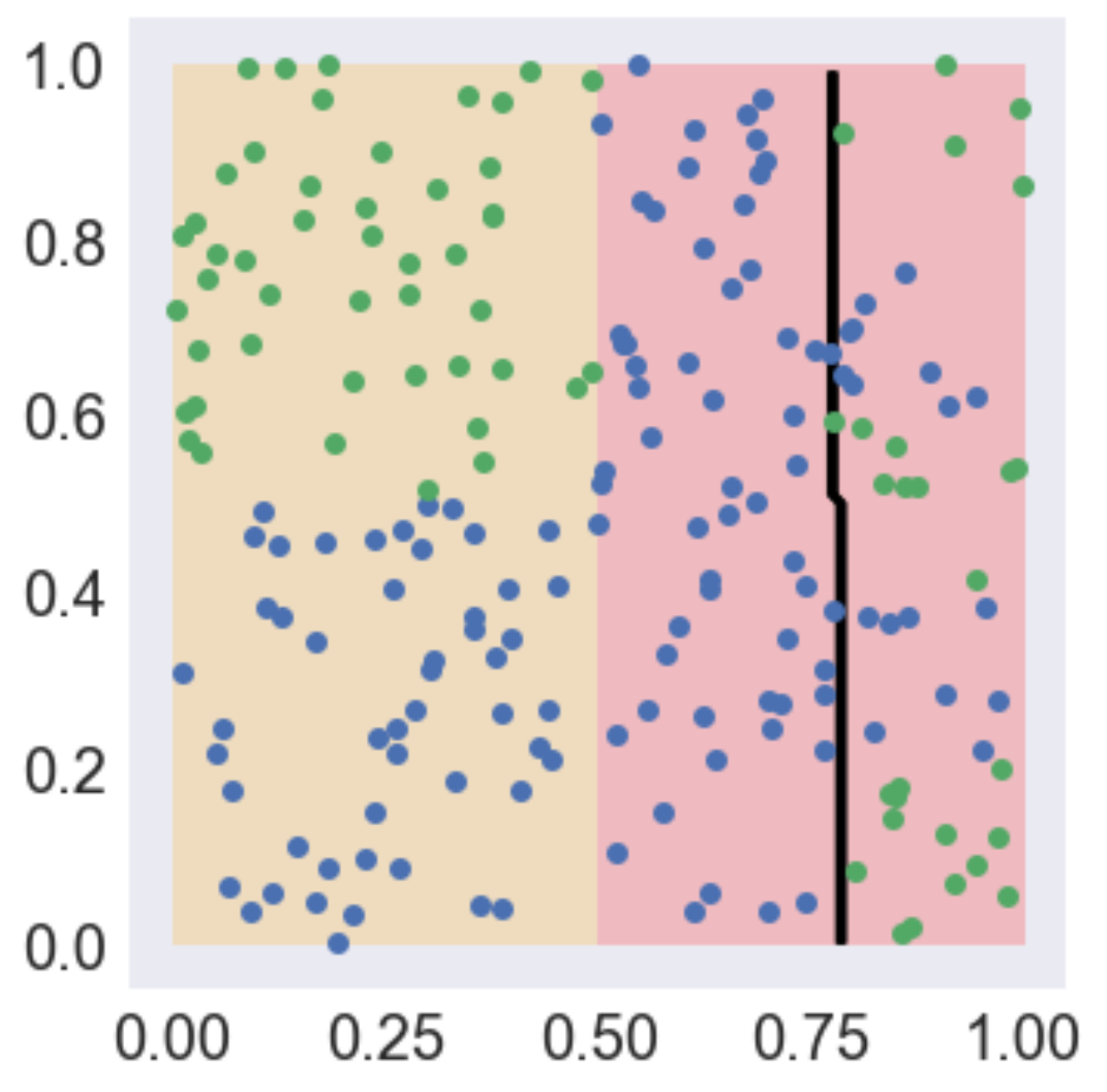}
    \caption{L$_1$ run B}
  \end{subfigure}
  \begin{subfigure}[b]{0.19\linewidth}
    \includegraphics[width=\linewidth]{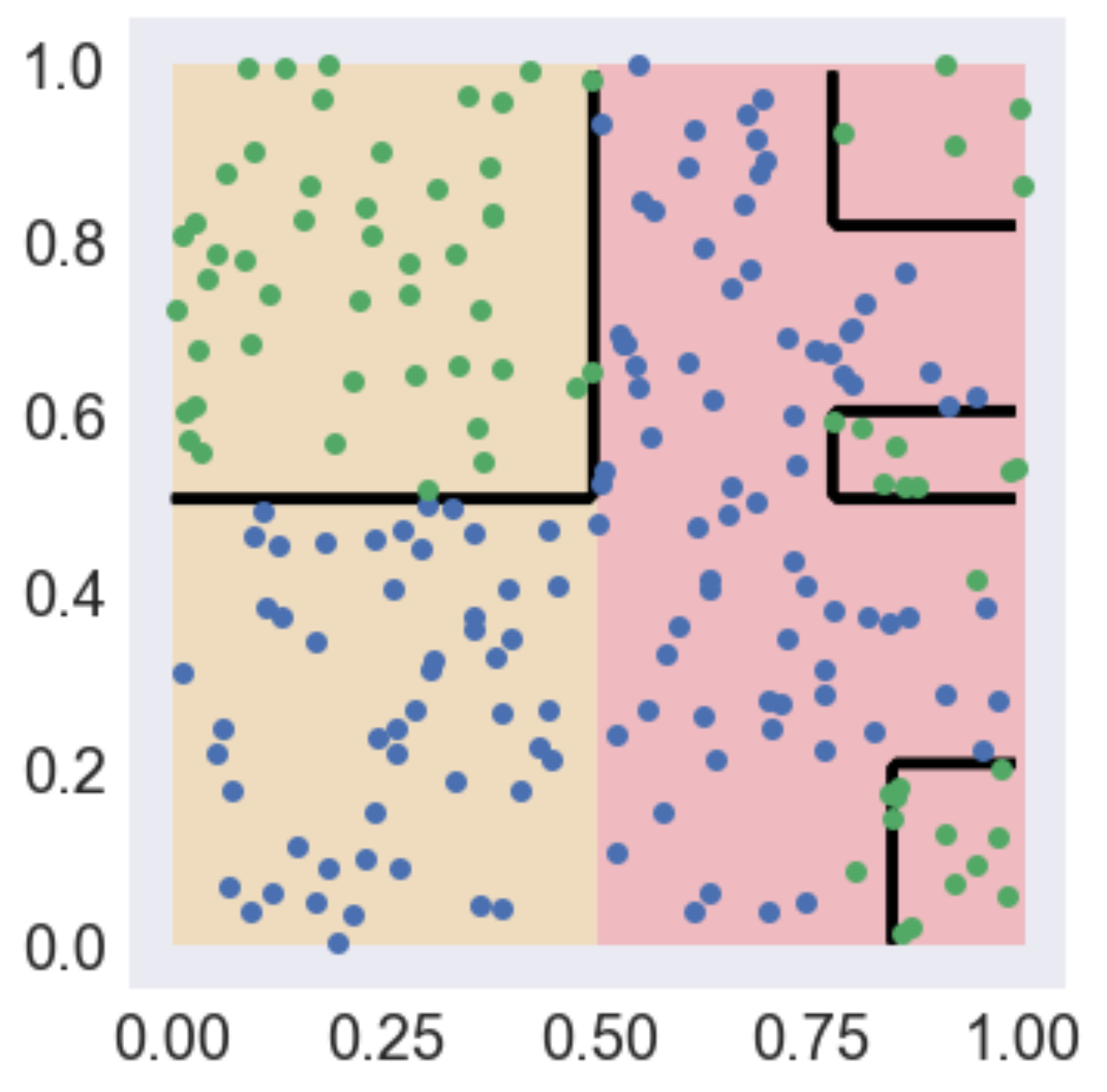}
    \caption{L$_0$ run A}
  \end{subfigure}
  \begin{subfigure}[b]{0.19\linewidth}
    \includegraphics[width=\linewidth]{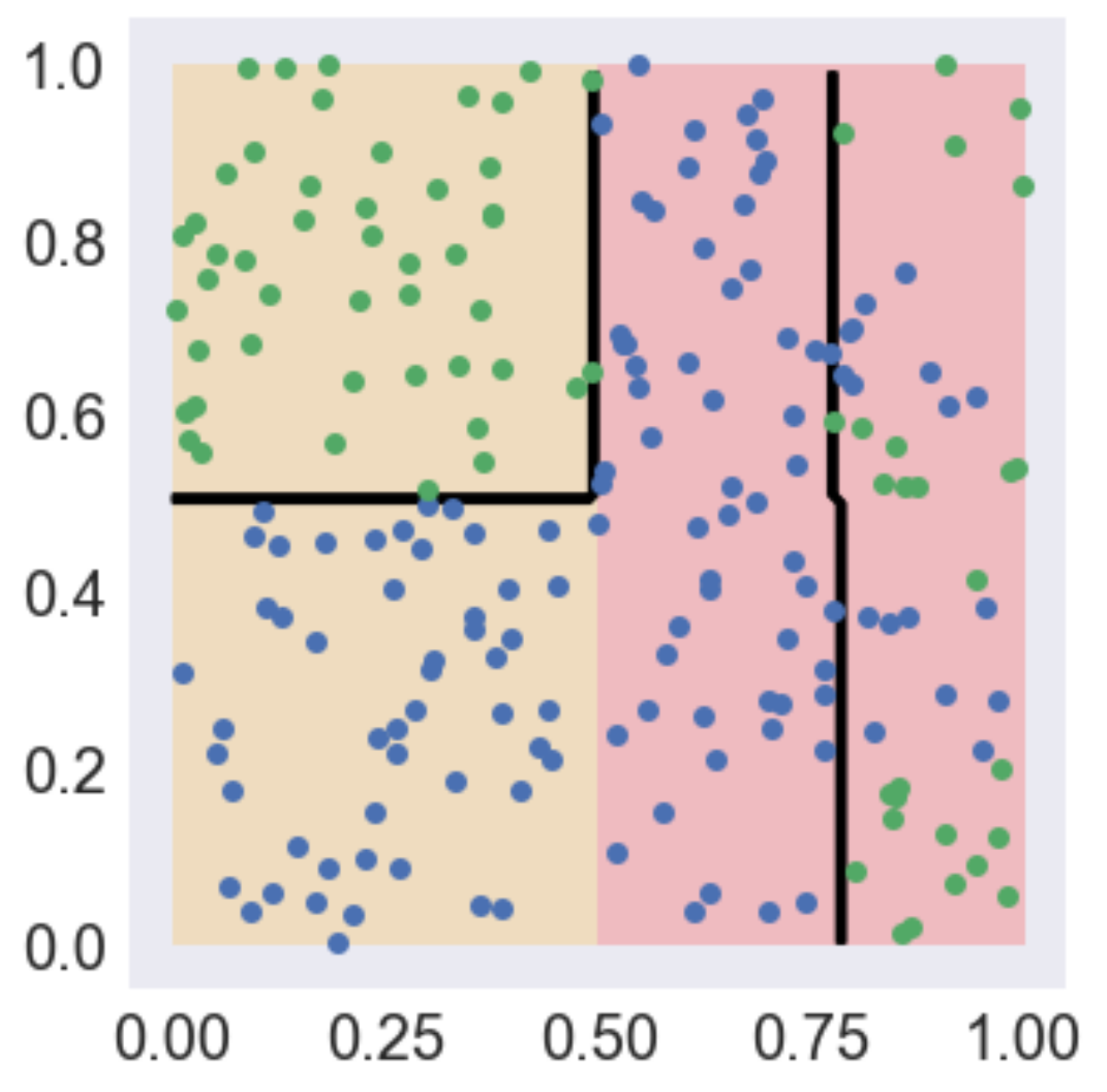}
    \caption{L$_0$ run B}
  \end{subfigure}
  \caption{
  Comparison of L$_1$ and L$_0$ penalties on per-region APL for blue-vs-green binary classification (each with two runs). Yellow and red patches represent regions. Empty regions denote ``trivial" decision functions.
}
  \label{fig:toyexp3}
  \vspace{-1em}
\end{figure}

To prevent regularization of simpler regions while other regions stay complex, we instead choose to penalize \emph{only} the average decision path length of the most complex region:
\begin{equation}
  \Omega^{\texttt{regional-L}0}(\theta)
  \triangleq
  \max_{r \in \{1, 2, \ldots R\}} \textsc{APL}(X_r, f(\cdot, \theta))
\label{eqn:argmax-apl}
\end{equation}
which corresponds to an L$_0$ norm over the path lengths, $\{\textsc{APL}(X_r, \cdot) \}_{r=1}^R$.  We will refer to this as L$_0$ \textit{regional tree regularization}.  In Fig.~\ref{fig:toyexp3}(d,e), we see this has signficant, desirable effects: L$_0$ regional tree regularization results in a more ``balanced" penalty between regions, leaving both regions with simple but nontrivial boundaries (d,e).

Eq.~\ref{eqn:argmax-apl} is differentiable as the gradient through a max operator masks all indexes but one. In practice, when training deep neural networks with the L$_0$ regional penalty in Eq.~\eqref{eqn:argmax-apl}, we face two technical difficulties:  first, the $\textsc{APL}$ subprocedure is not differentiable; second, regularizing only one region at a time significantly slows down convergence. Both represent key challenges we overcome with technical contributions detailed in the next two subsections.

\subsection{Innovation: Use SparseMax Penalty across Regions}

While it solves the over-regularization problem we faced with the L$_1$ regional penalty, the $\texttt{max}$ in the L$_0$ regional penalty forces us to regularize only one region at a time. In our experiments we found this to cause L$_0$ regularized models to train for a much longer time before converging to a minima. For example, on the UCI datasets, holding all hyperparameters and architectures fixed, an L$_0$ regularized deep network took around 900 epochs to converge, 9 times longer than other regularizations (convergence is measured by APL and accuracy on a validation set that does not change for at least 10 epochs). We believe this to be due to oscillatory behavior where two regions take turns (1) growing more complex in order to improve accuracy and (2) growing less complex to in order to reduce the regularization cost. Because Eq.~\eqref{eqn:argmax-apl} only ever regularizes a single region at a time, this oscillation can prolong training.

This is not ideal as we would like our regularizers to not introduce computational cost over unregularized models. The oscillatory behavior of the $\texttt{max}$ in Eq.~\eqref{eqn:argmax-apl} is a challenge new to our region-specific approach (this is not observed in  global tree regularization). Naively, we would solve this problem by increasing the number of regions we regularize at once. However,
common approximations to $\texttt{max}$, like $\texttt{softmax}$, are not sparse and include non-zero contributions from all regions.
This makes it difficult to focus on the most complex regions as $\max$ does. In experiments with $\texttt{softmax}$, we observed the same problematic behavior that L$_1$ penalties exhibited: it tended to make some regional boundaries far too simple.

To balance two competing interests, we apply the recently-proposed $\textsc{SparseMax}$ transformation~\citep{martins2016softmax},
which can attend solely to the most problematic regions (setting others to zero contribution) while remaining differentiable (a.e.). Intuitively, $\textsc{SparseMax}$ corresponds to a Euclidean projection of an length-$R$ vector of reals (in our case, one APL value per region) to a length-$R$ vector of non-negative entries that sums to one.
When the projection lands on a boundary in the simplex, then the resulting probability vector will be sparse.  Our chosen $\textsc{SparseMax}$ approximation means that we penalize only a few of the most complex regions, avoiding over-regularization. Also, because $\textsc{SparseMax}$ often regularizes more than one region at a time, it avoids oscillation and takes a comparable number of epochs as other regularizers to converge. Alg.~\ref{algorithm:3} details \textsc{SparseMax} applied to our regularizer, named L$_{\text{SP}}$ \textit{regional tree regularization}.

\begin{algorithm}[H]
\caption{\textsc{L$_{\text{SP}}$ Regional Tree Reg.}}
\begin{algorithmic}[1]
\Require{
  \Statex $\mathbf{\hat{\Omega}} = \{\hat{\Omega}^{\texttt{regional}}_r\}_{r=1}^R$: APL for each of $R$ regions
}
\Function{$\textsc{SparseMax}$}{$\mathbf{\hat{\Omega}}$}
   \State Sort $\mathbf{\hat{\Omega}}$ such that $\mathbf{\hat{\Omega}}[i] \geq \mathbf{\hat{\Omega}}[j]$ if $i \geq j$
   \State $k = \max \{ r \in [1, R] | (1 + r\mathbf{\hat{\Omega}}[r]) > \sum_{i \leq r} \mathbf{\hat{\Omega}}[i] \}$
   \State $\tau = k^{-1}(-1 + \sum_{i \leq k} \mathbf{\hat{\Omega}}[i])$
   \State \Return $\{p_r\}_{r=1}^R$ where $p_r = \max\{\mathbf{\hat{\Omega}}_r - \tau, 0 \}$
\EndFunction
\end{algorithmic}
\label{algorithm:3}
\end{algorithm}





\subsection{Innovation: Three Keys to Reliable Optimization}
\label{sec:global:improve}

The non-differentiability of the \textsc{APL} subroutine in Eq.~\eqref{eqn:argmax-apl} can be addressed by training a surrogate estimator of APL for each region. In the Appendix, we extend the  training procedure from Eq.~\eqref{eqn:surrogate} to the region-specific case.

Optimizing surrogate networks is a delicate operation. Even when training only one surrogate for global tree regularization, as in \citet{wu2018beyond}, we found that  the surrogate's ability to accurately predict the APL was very sensitive to chosen hyperparameters such as learning rates.
Repeated runs from different random initializations also often found different minima---making tree regularization unreliable. These issues were only exacerbated when training multiple surrogates, where we are trying to train many estimators that each focus on smaller regional datasets. We found that sophistication is needed to keep the gradients accurate and variances low.  Below, we list several optimization innovations that proved to be key to stabilizing training.

\begin{table}
\centering
\small
\caption{Comparison of the average and max mean squared error (MSE) between surrogate predictions and true APLs over five runs of 500 epochs. A lower MSE is desirable.
\vspace{-1em}
}
\begin{tabular}{l c c}
\toprule
Experiment & Mean MSE & Max MSE \\
\midrule
No data augmentation & $0.069 \pm 0.008$ & $0.987 \pm 0.030$\\
With data augmentation & $0.015 \pm 0.005$ & $0.298 \pm 0.017$ \\
\hline
Non-Deterministic Training & $0.116 \pm 0.011$ & $1.731 \pm 0.021$ \\
Deterministic Training & $0.024 \pm 0.006$ & $0.371 \pm 0.040$ \\
\bottomrule
\end{tabular}
\label{table:tricks}
\end{table}

\paragraph{Key 1. Data augmentation}
Small changes in the target model can make large changes to the APL for a specific region.  As such, regional surrogates need to be retrained frequently.  The practice from \cite{wu2018beyond} of computing the true APL for a dataset $\mathcal{D}^\theta$ gathered from $\theta$ values seen over  recent gradient descent iterations did not compile a large enough dataset to generalize well to new parameters $\theta'$.
Thus, we supplement the dataset with 100-1000 randomly sampled weight vectors: given the $J$ previous vectors $\{\theta \}_{j=1}^J$ stored in $\mathcal{D}^\theta$,
we form a new ``synthetic'' parameter vector $\theta'$ as a convex combination mixing weights drawn from a $J$-dimensional Dirichlet distribution with $\alpha_j = 1$ for $j=1, ..., J$. Each synthetic vector is paired with its true APL value. Table~\ref{table:tricks} how this reduces noise in predictions.

\begin{figure}[h!]
  \centering
  \begin{subfigure}[b]{0.19\linewidth}
    \includegraphics[width=\linewidth]{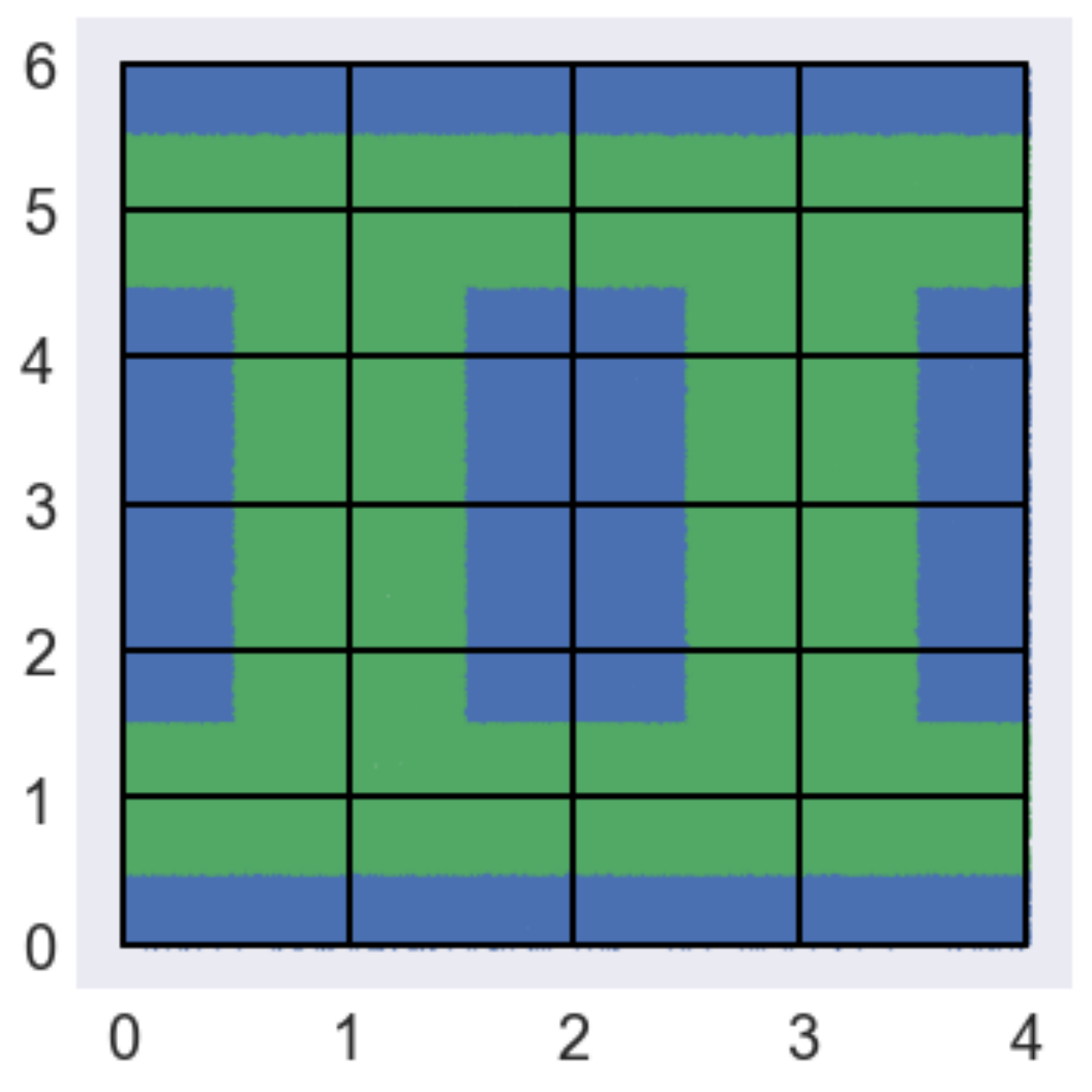}
    \caption{}
  \end{subfigure}
  \begin{subfigure}[b]{0.19\linewidth}
    \includegraphics[width=\linewidth]{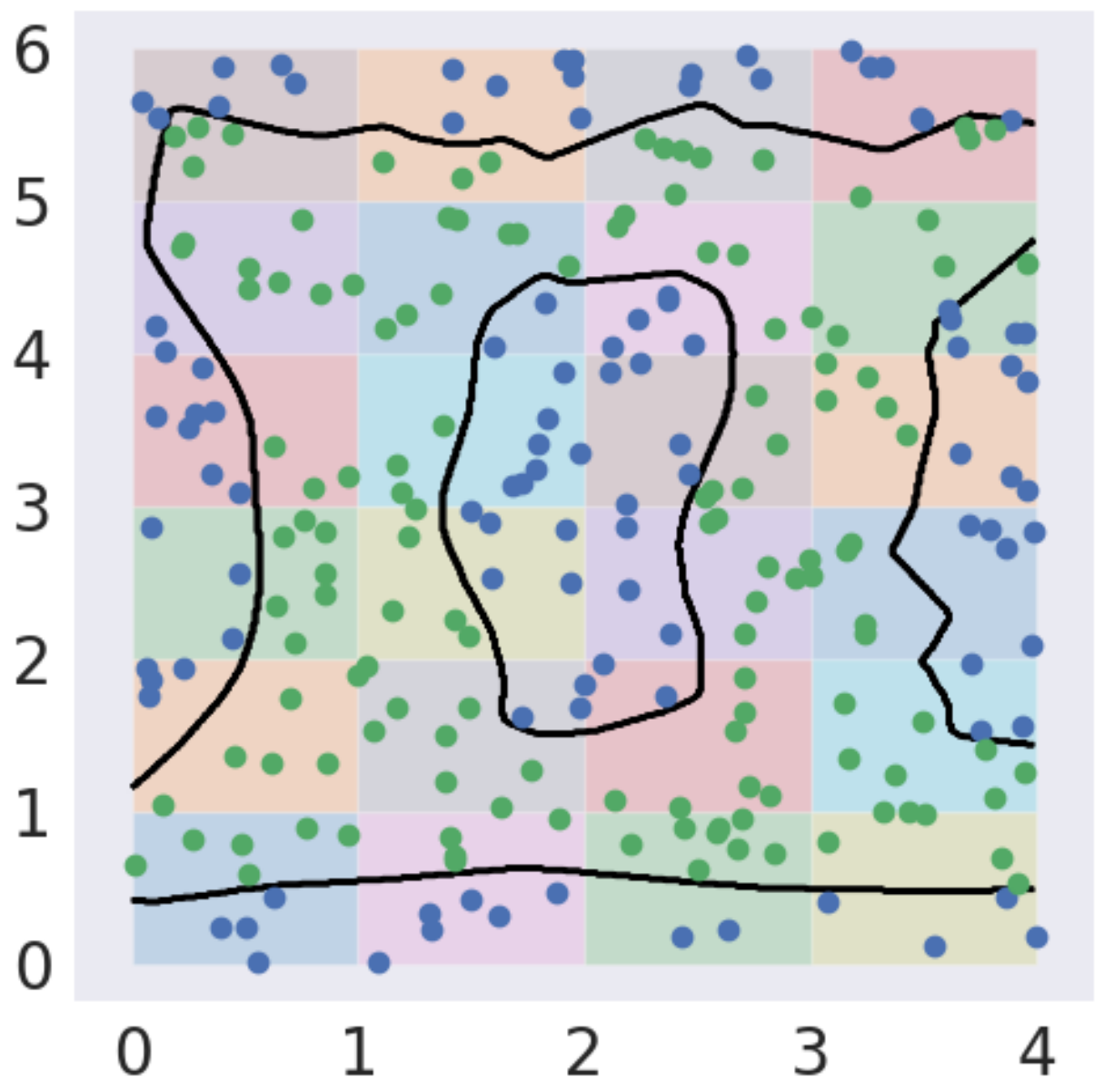}
    \caption{}
  \end{subfigure}
  \begin{subfigure}[b]{0.19\linewidth}
    \includegraphics[width=\linewidth]{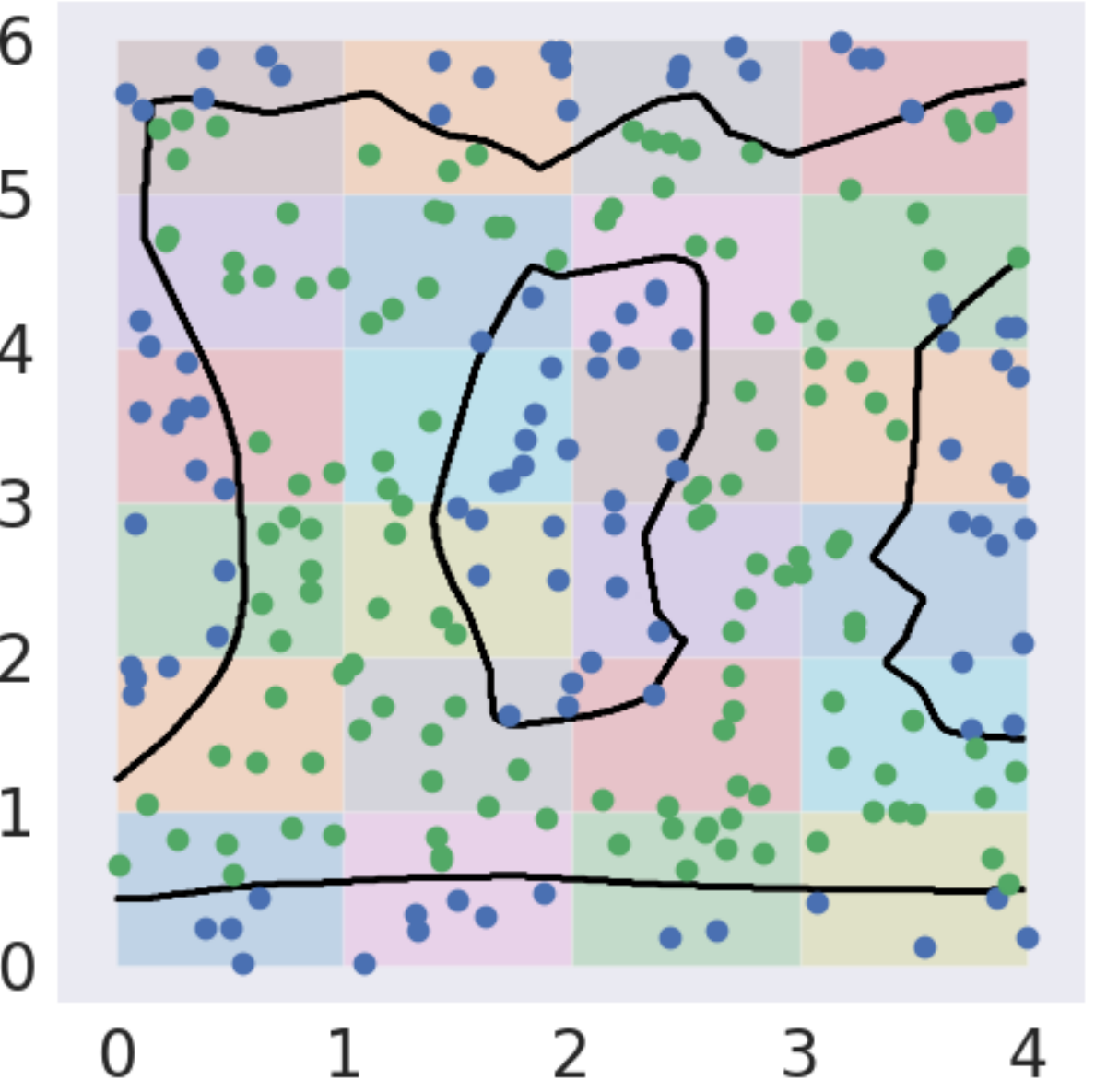}
    \caption{}
  \end{subfigure}
  \begin{subfigure}[b]{0.19\linewidth}
    \includegraphics[width=\linewidth]{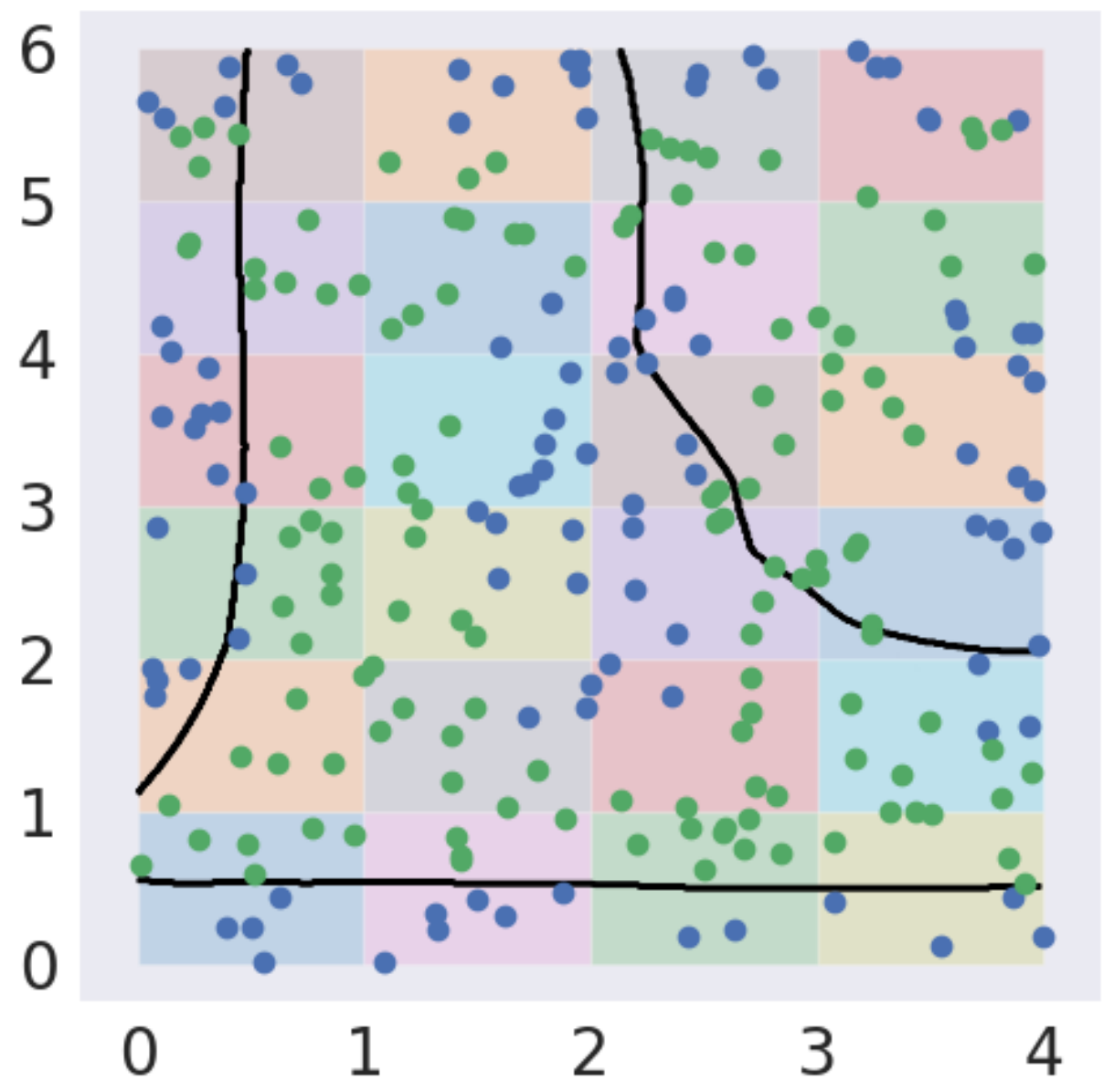}
    \caption{}
  \end{subfigure}
  \begin{subfigure}[b]{0.19\linewidth}
    \includegraphics[width=\linewidth]{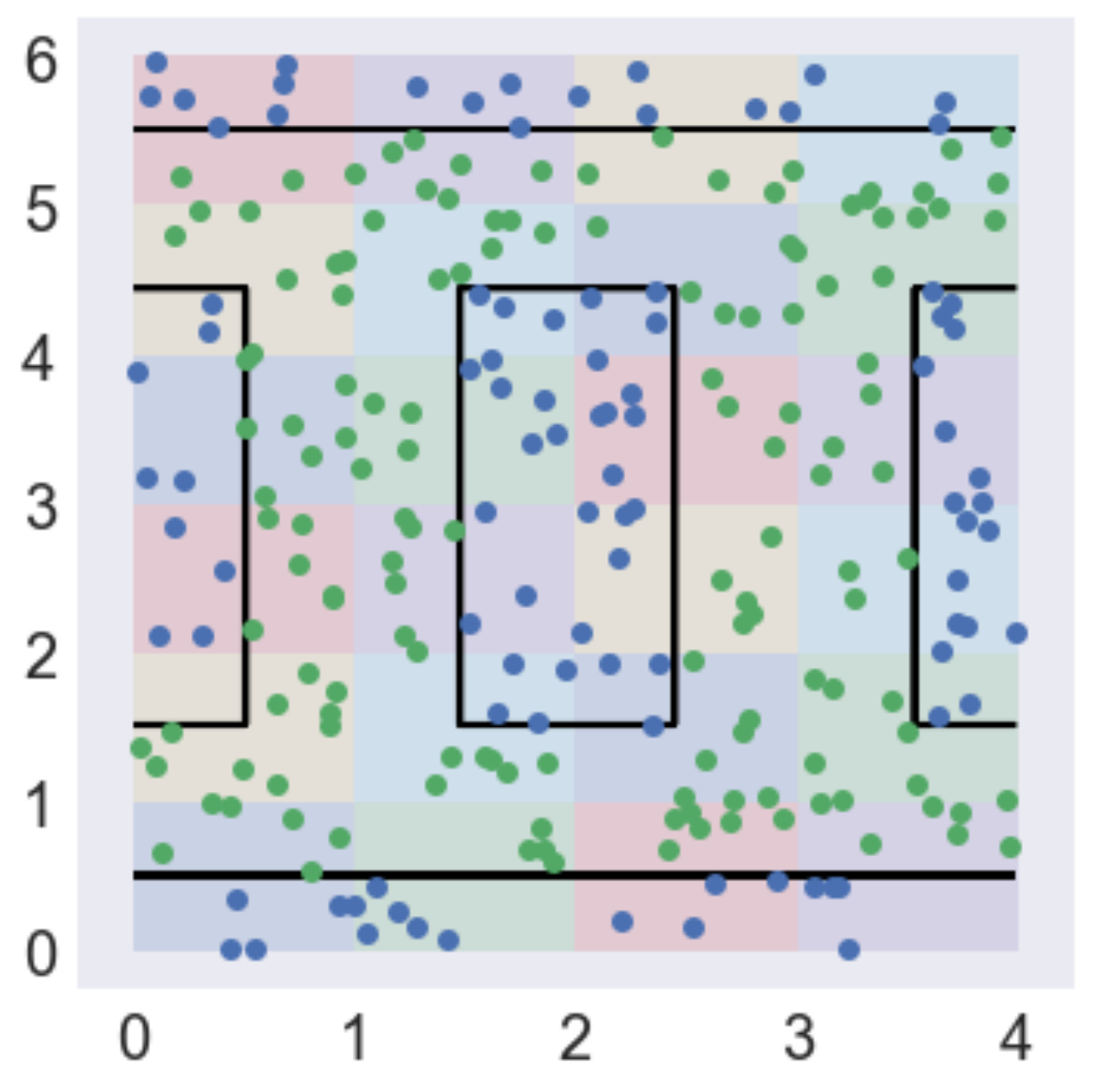}
    \caption{}
  \end{subfigure}
  \caption{
  Ablation analysis on a toy green-vs-blue classification task. Colored patches represent regions.
  (a) Ground truth labels. (b) Minima with no regularization. (c) Minima with no data augmentation. (d) Minima with no pruning or determinism in training trees. (e) Minima using  innovations. }
  \label{fig:toyexp4}
  \vspace{-1.5em}
\end{figure}

\paragraph{Key 2. Deterministic CART.}
CART is a common algorithm to train decision trees. In short, CART enumerates over all unique values for every input feature and computes the Gini impurity as a cost function; it chooses the best split as the one with the lowest impurity. To make computation more affordable, implementations of CART, such as the one we use in Scikit-Learn \citep{pedregosa2011scikit}, randomly sample a subset of features enumerate over (rather than all features). As such, multiple runs will result in different trees of different APL (see Appendix). Unexplained variance in CART makes fitting a surrogate more difficulty -- in fact, this variance compounds over multiple surrogates. Fixing the random seed eliminates this issue and results in better predictions (see Table~\ref{table:tricks}). Alternatively, to ensure that CART is not choosing the same subset of features over and over (which may be biased), one can choose a fixed set of random seeds and compute the true APL as the average over several CART runs, one for each of the seeds.

\begin{figure*}[t!]
  \centering
  \begin{subfigure}[b]{0.13\textwidth}
    \includegraphics[width=\textwidth]{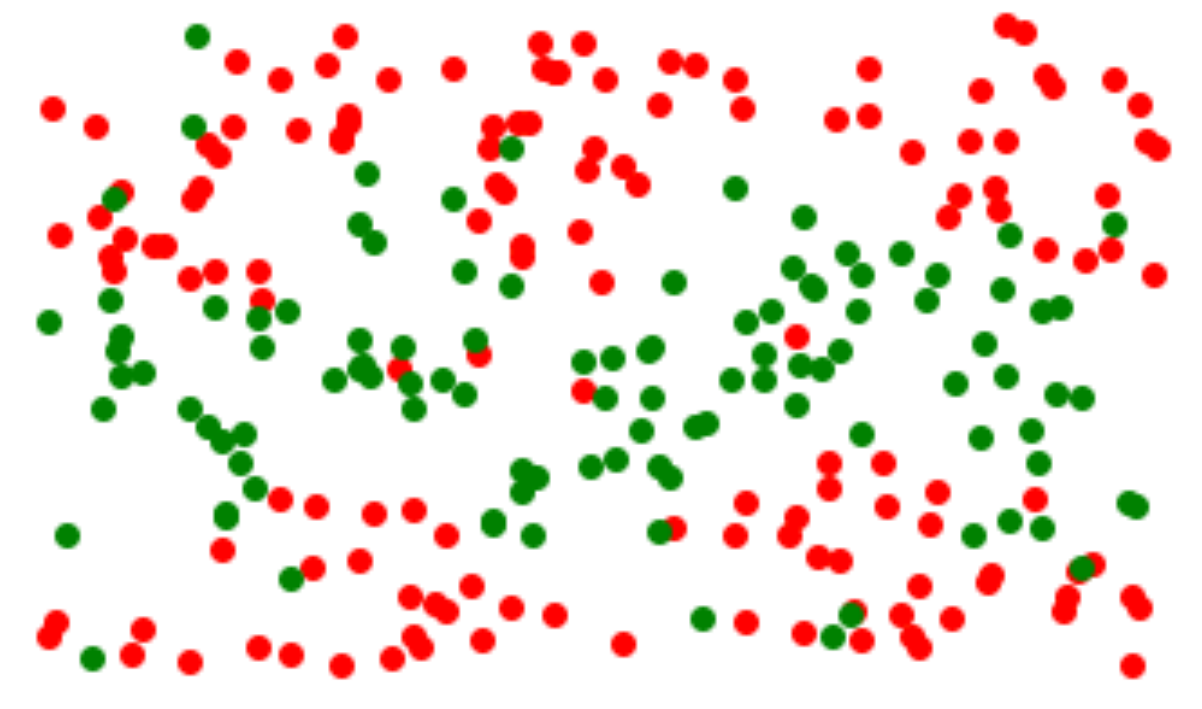}
    \caption{Training set}
  \end{subfigure}
  \begin{subfigure}[b]{0.13\textwidth}
    \includegraphics[width=\textwidth]{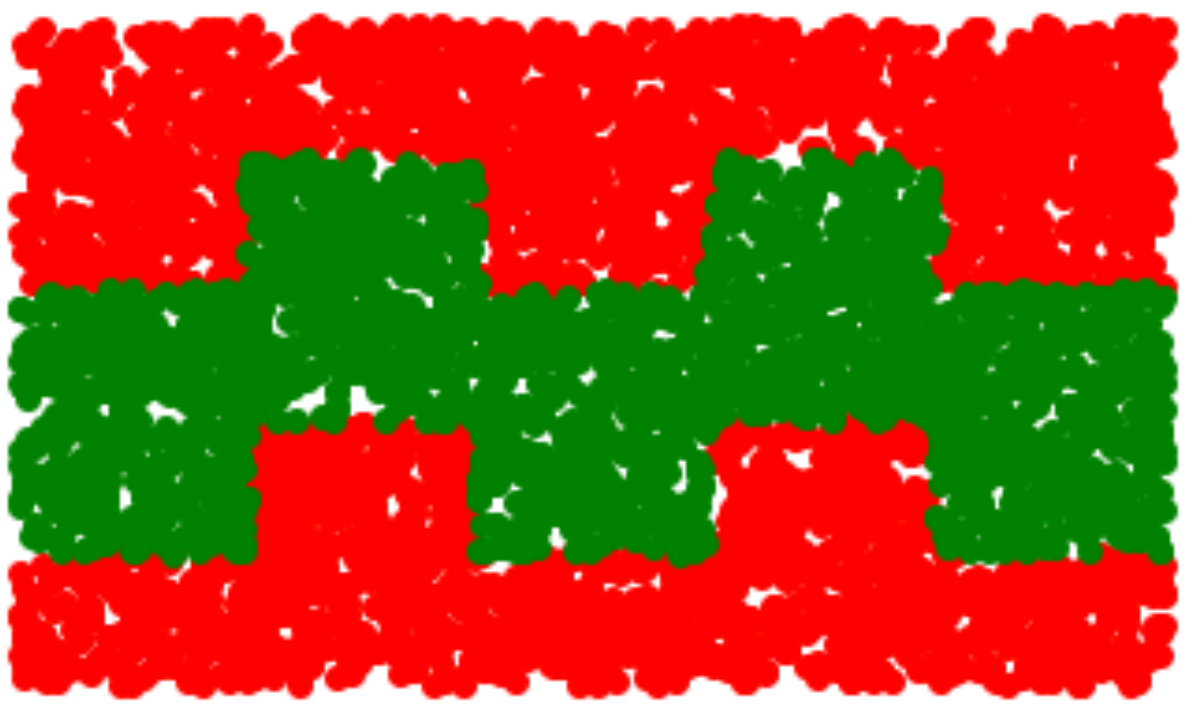}
    \caption{Test set}
  \end{subfigure}
  \begin{subfigure}[b]{0.13\textwidth}
    \includegraphics[width=\textwidth]{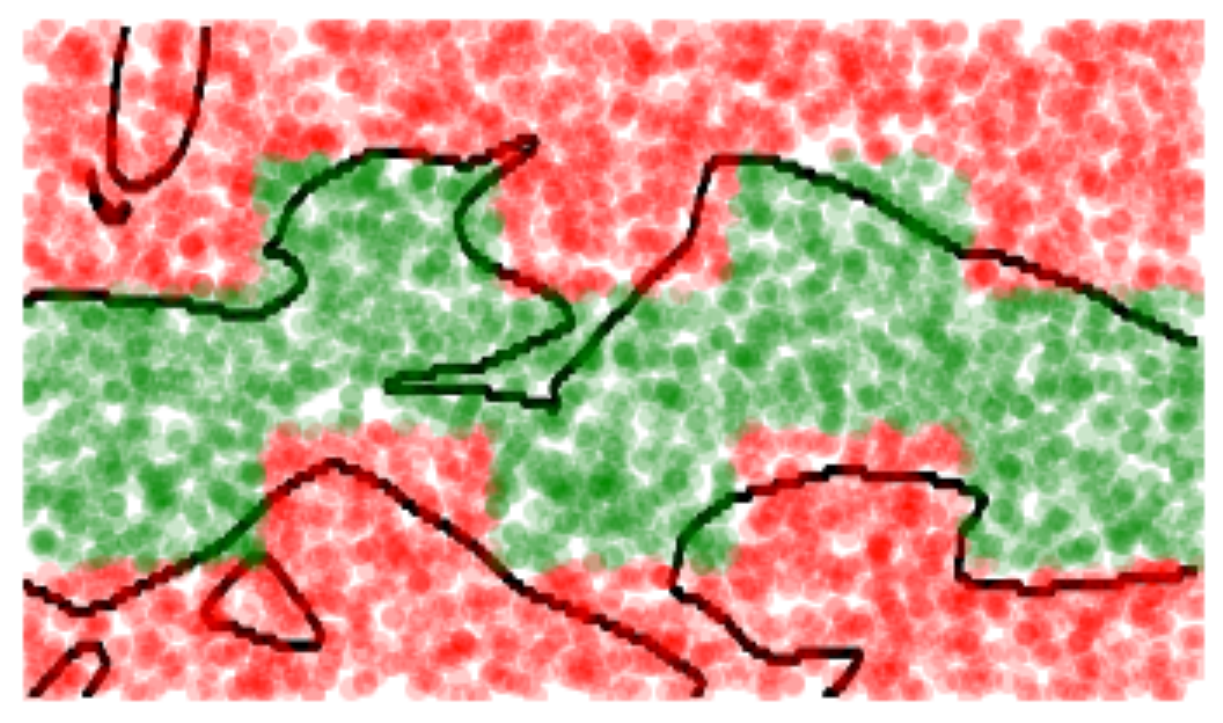}
    \caption{Unregularized}
  \end{subfigure}
  \begin{subfigure}[b]{0.13\textwidth}
    \includegraphics[width=\textwidth]{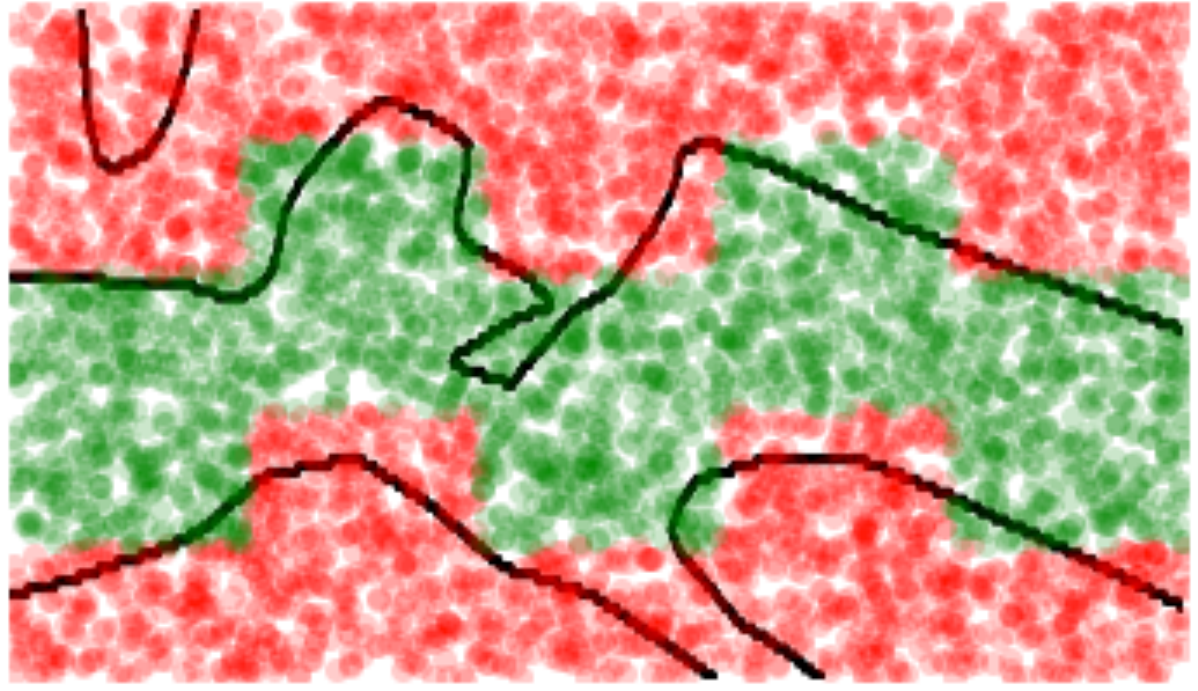}
    \caption{L$_2$}
  \end{subfigure}
  \begin{subfigure}[b]{0.13\textwidth}
    \includegraphics[width=\textwidth]{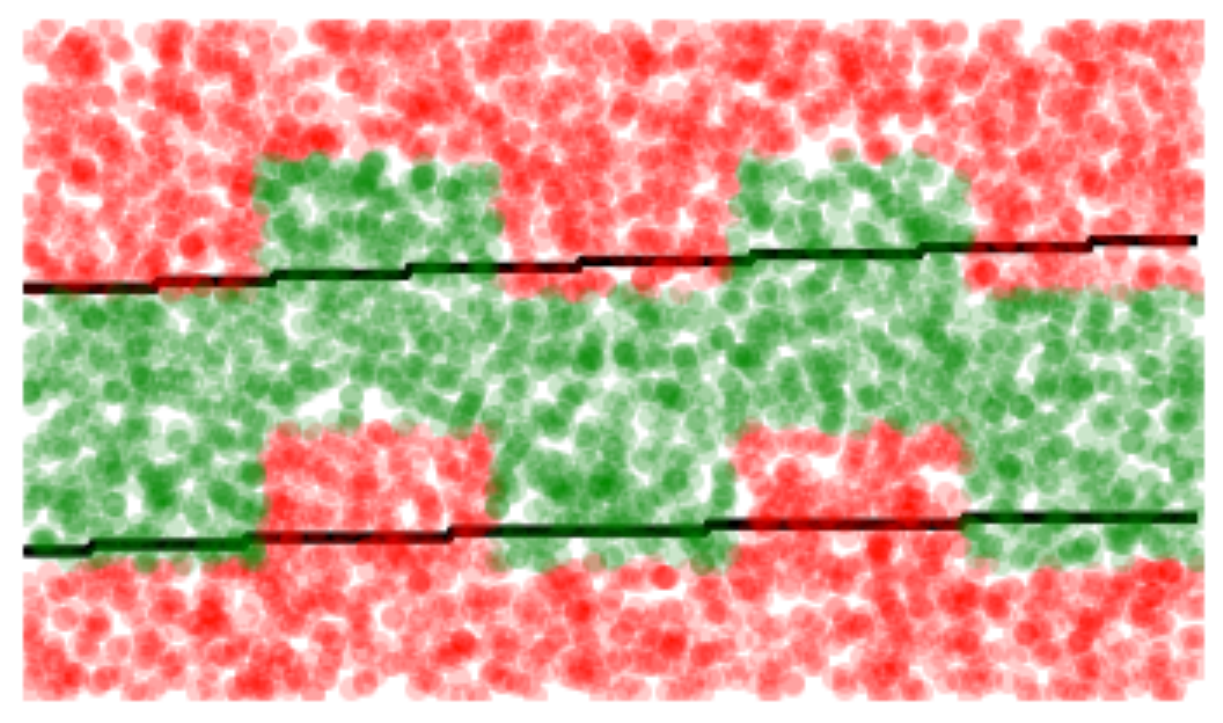}
    \caption{Global Tree}
  \end{subfigure}
  \begin{subfigure}[b]{0.13\textwidth}
    \includegraphics[width=\textwidth]{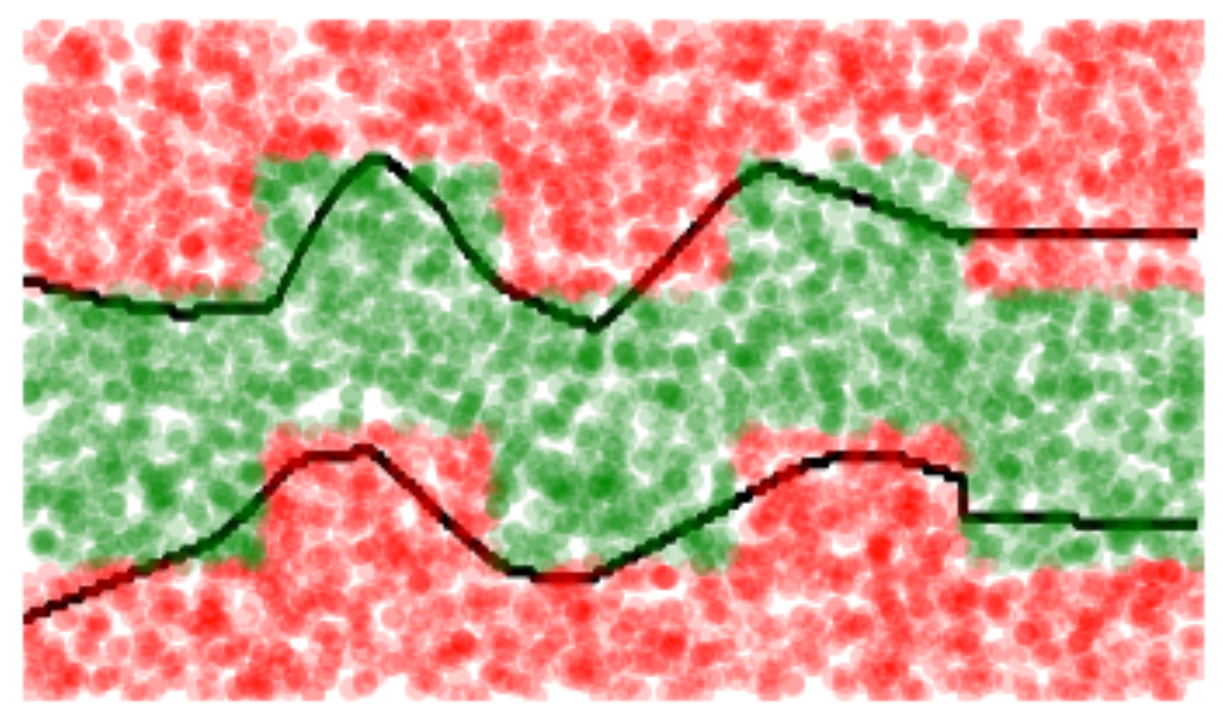}
    \caption{L$_1$ Reg. Tree}
  \end{subfigure}
  \begin{subfigure}[b]{0.13\textwidth}
    \includegraphics[width=\textwidth]{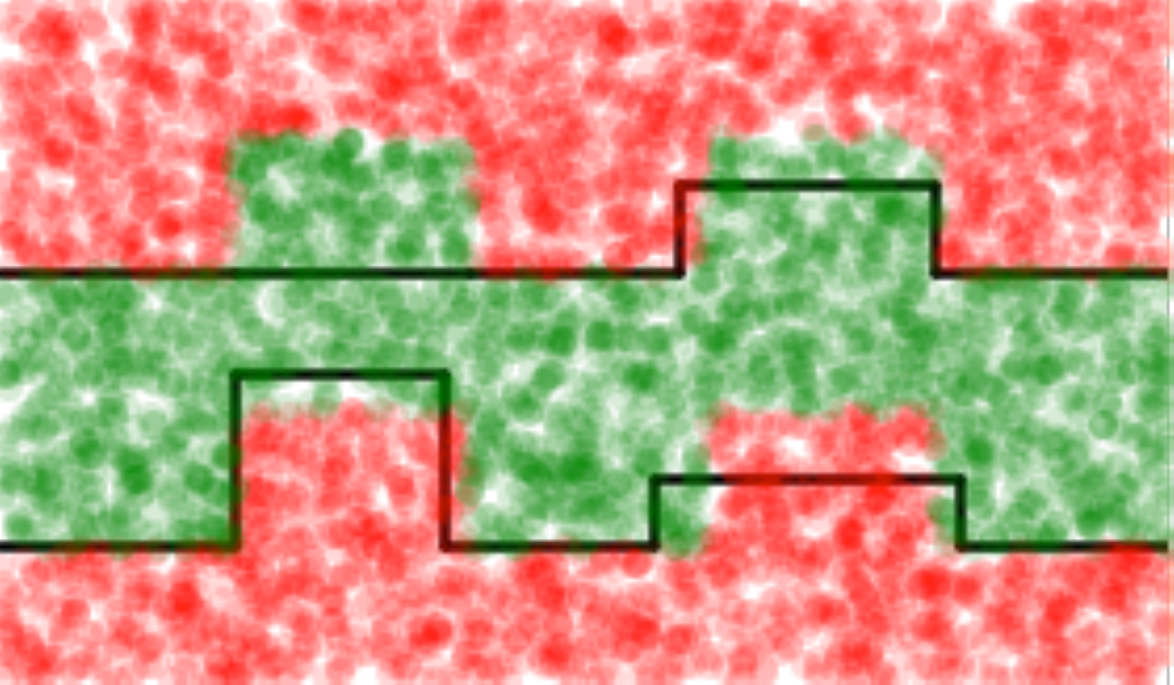}
    \caption{L$_\text{SP}$ Reg. Tree}
  \end{subfigure}
  \caption{Synthetic data with a sparse training set (a) and a dense test set (b). Due to sparsity, the division of five rectangles is not trivial to uncover from (a). (c-g) show contours of decision functions learned with varying regularizations and strengths.}
  \label{fig:toy}
\end{figure*}

\paragraph{Key 3. Pruning decision trees}
Given a dataset $\mathcal{D}$ of feature vectors and class labels, even with a fixed seed there are many possible decision trees that can fit $\mathcal{D}$ with equal accuracy. One can always add additional subtrees that predict the same label as the parent node, thereby keeping accuracy constant but adding to the tree's depth (and thus APL). This again introduces difficulty in learning a surrogate.

To overcome this, we add a \textsc{PruneTree} post-processing step to the APL computation in Alg.~\ref{algorithm:2}.
We use \textit{reduced error pruning}~\citep{quinlan1987simplifying}, which removes any subtree that does not effect performance as measured on a validation dataset not used in $\textsc{TrainTree}$. See Alg.~\ref{algorithm:3} in the Appendix for an updated algorithm.
We emphasize that pruning dramatically improves the stability of results.

\paragraph{Ablation analysis.}
Together, these innovations were  crucial for learning with regional tree regularization. Fig.~\ref{fig:toyexp4} shows an ablation study on a synthetic green-vs-blue classification task with 25 regions.
Without data augmentation (c), there are not enough examples to fully train each surrogate, so the tree regularization penalty is effectively non-existant and the model finds similar minima to no regularization (b).
Without pruning and fixing seeds (d), the APLs vary due to randomness in fitting a decision tree. This leads to strange and ineffective decision boundaries. Only with all innovations (f) do we converge to an accurate decision boundary that remains simulable in each region.

\section{Experiments}


\paragraph{Evaluation Metrics}
We wish to compare models with global and regional explanations. However, regional and global APL are not directly comparable: subtly, the APL of a global tree is an overestimate in a single region. To reconcile this, for any globally regularized model, we separately compute $\Omega^{\texttt{regional-L}1}$ as an evaluation criterion. In this context, $\Omega^{\texttt{regional-L}1}$ is used only for evaluation; it does not appear in the objective. We do the same for all baseline models. From this point on, if we refer to APL (e.g. Test APL, average path length) outside of training, we are referring to the evaluation metric, $\Omega^{\texttt{regional-L}1}$. For classification error, we measure either F1 score or AUC on a held out test set.

\paragraph{Baselines}

For each dataset, we compare our proposed L$_\text{SP}$ regional tree regularization to several alternatives.
First, we consider other tree regularization approaches for MLPs, including
global tree regularization \citep{wu2018beyond} (to show the benefit of regional decomposition),
and L$_1$ regional tree regularization (to show why the L$_\text{SP}$ penalty in Eq.~\eqref{eqn:argmax-apl} is needed). Second, we compare to $L_0$ regional tree regularization (to show that L$_\text{SP}$ finds similar minima, only faster).
Third, we consider simpler ways of training MLPs, such as no regularization and an L$_2$ penalty on parameters.
Finally, we need to demonstrate the benefits of regularizing neural networks to be interpretable rather than simply training standalone decision trees directly.
We thus compare to a global decision tree classifier and an ensemble of tree classifiers (one for each region). We call these two ``Decision Tree'' and ``Regional Decision Tree''.

\paragraph{Model Selection}
We train each regularizer with an exhaustive set of strengths: $\lambda$ = 0.0001, 0.0005, 0.001, 0.005, 0.01, 0.02, 0.05, 0.1, 0.2, 0.5, 1.0, 2.0, 5.0, 10.0. Three runs with different random seeds were used to avoid local optima.
\begin{figure*}[h!]
  \centering
  \begin{subfigure}[b]{\textwidth}
    \includegraphics[width=\textwidth]{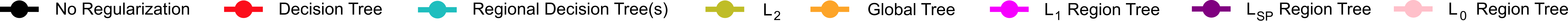}
  \end{subfigure}
  \begin{subfigure}[b]{0.24\textwidth}
    \includegraphics[width=\textwidth]{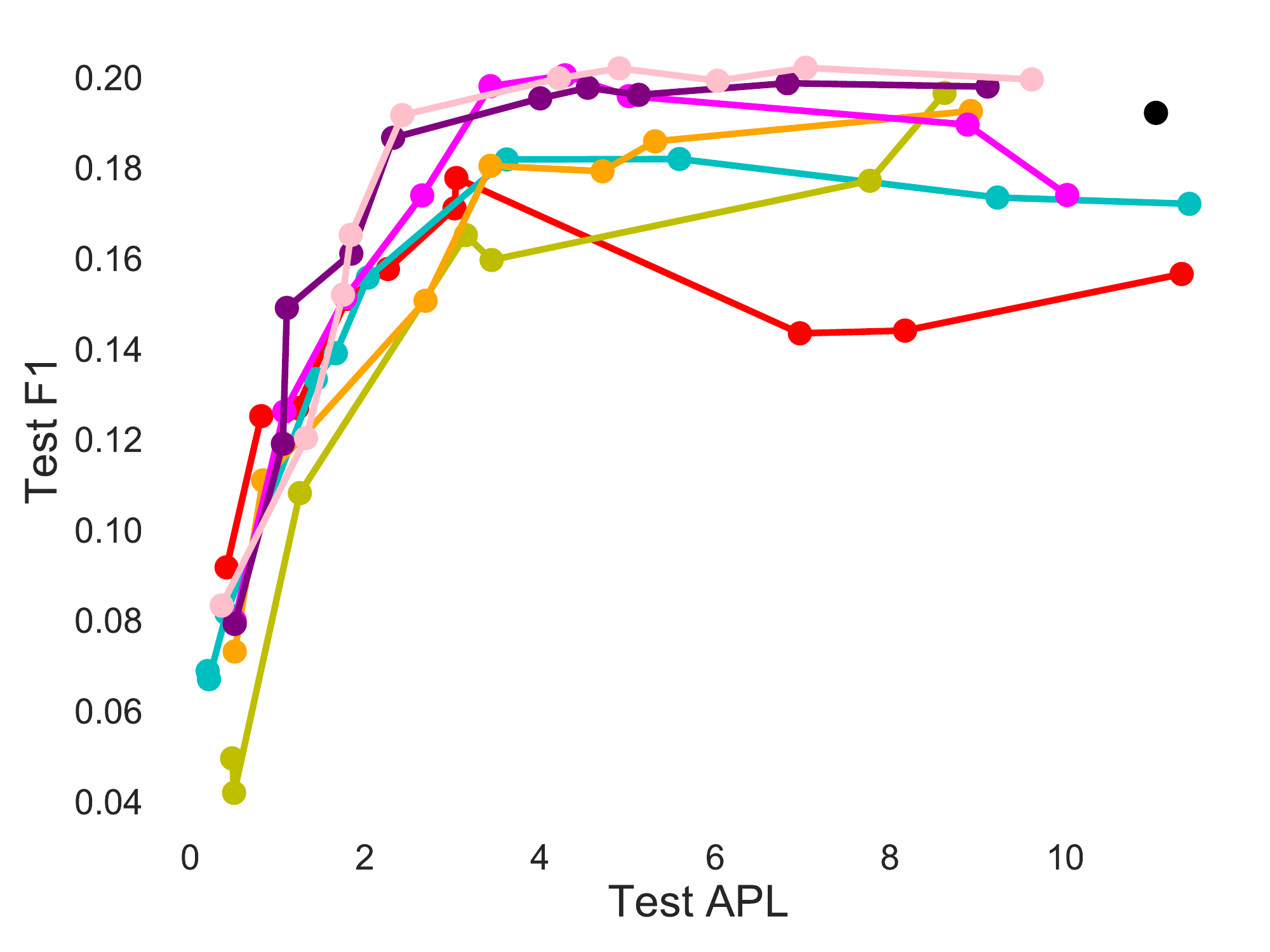}
    \caption{Bank}
  \end{subfigure}
  \begin{subfigure}[b]{0.24\textwidth}
    \includegraphics[width=\textwidth]{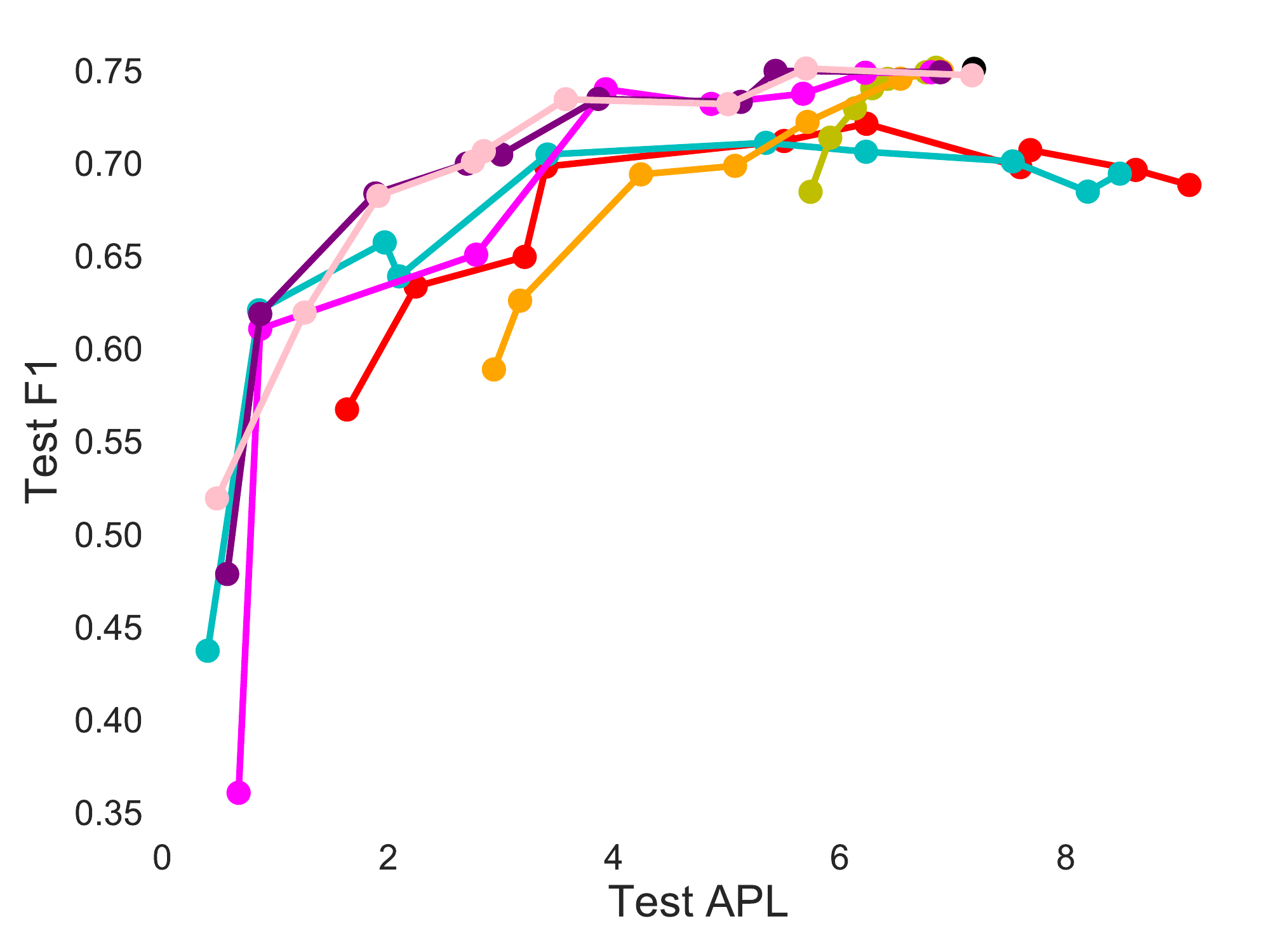}
    \caption{Gamma}
  \end{subfigure}
  \begin{subfigure}[b]{0.24\textwidth}
    \includegraphics[width=\textwidth]{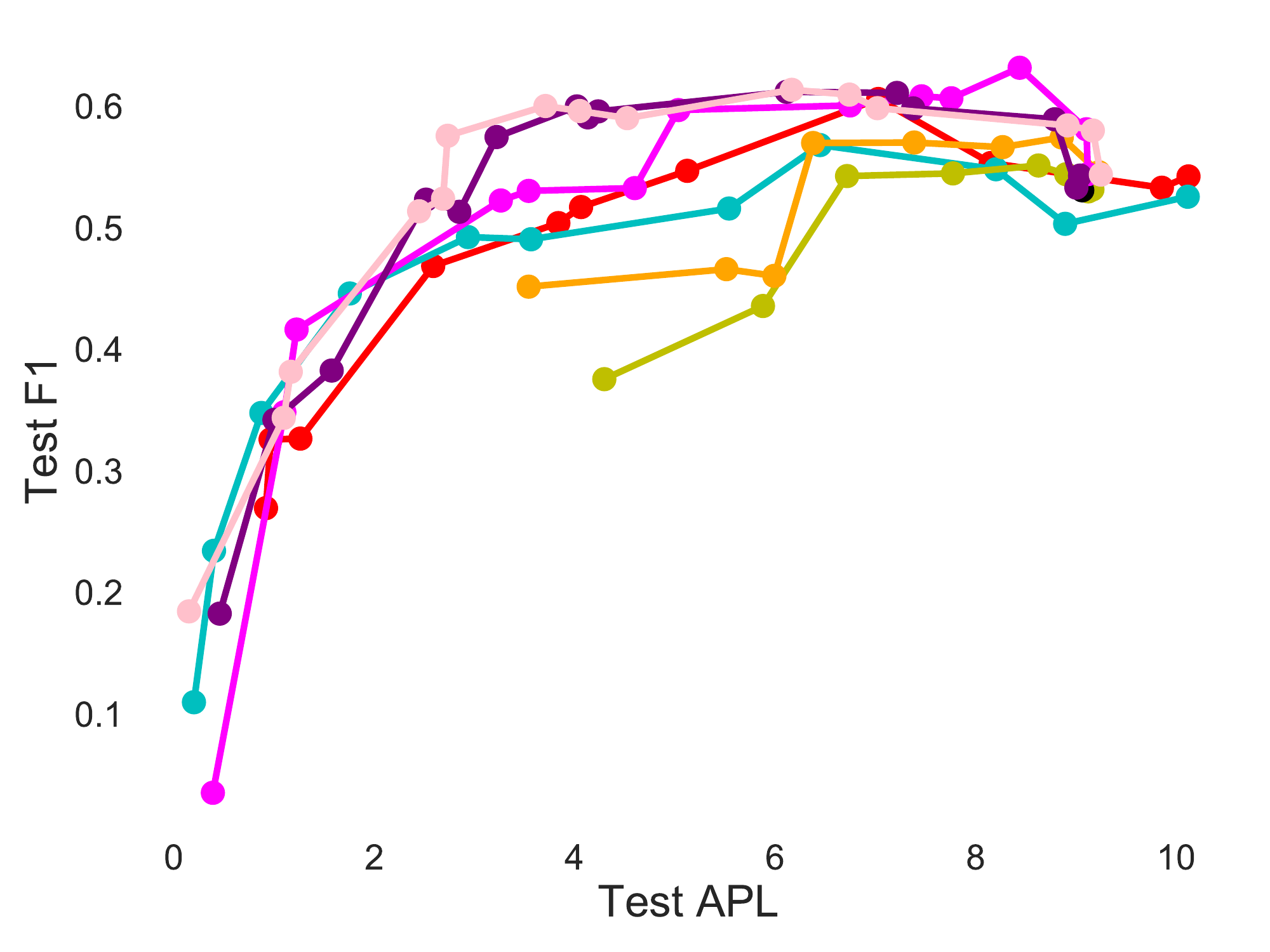}
    \caption{Adult}
  \end{subfigure}
  \begin{subfigure}[b]{0.24\textwidth}
    \includegraphics[width=\textwidth]{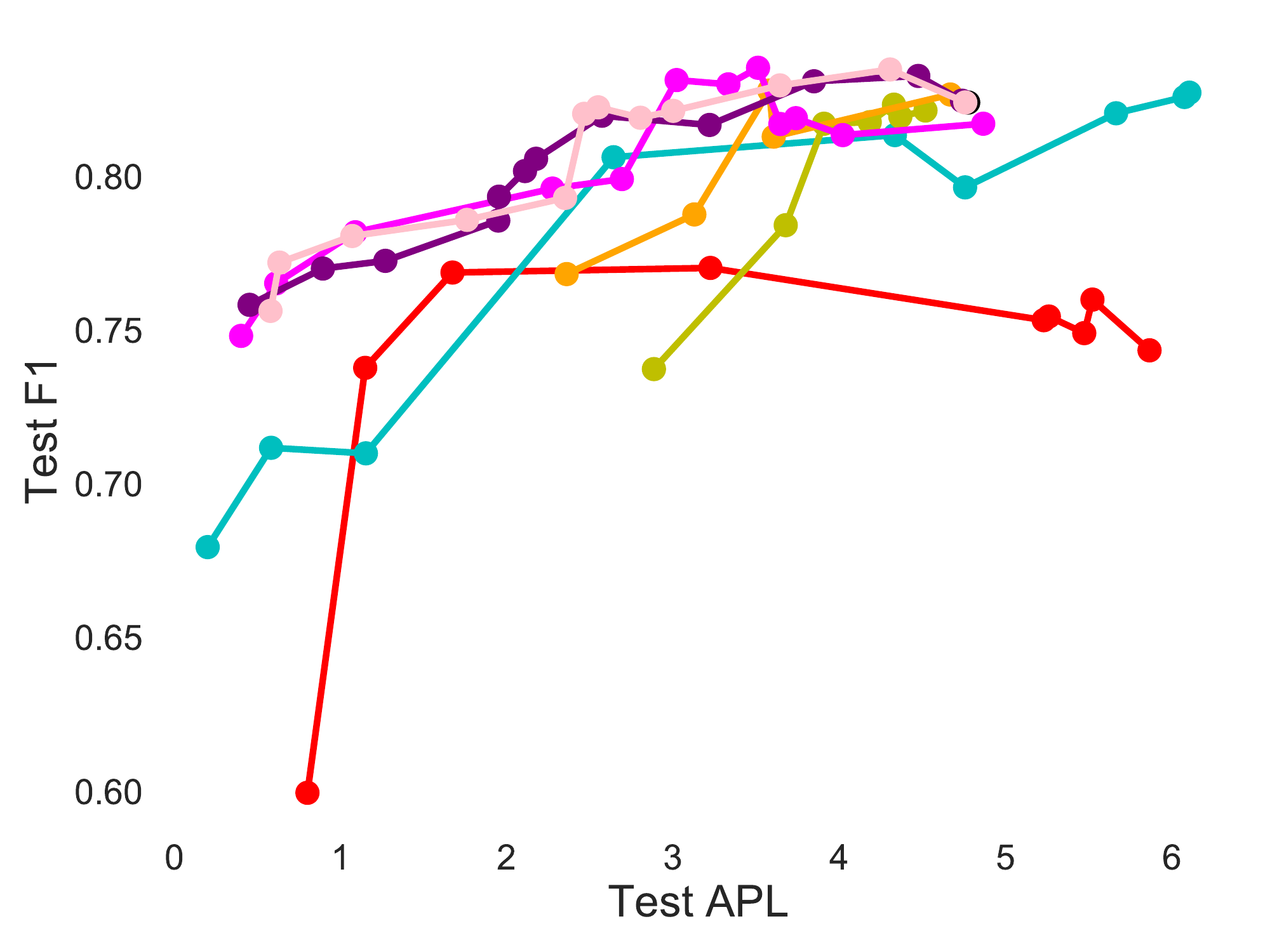}
    \caption{Wine}
  \end{subfigure}
  \caption{
  Prediction quality vs. simulability tradeoff curves on four UCI datasets.
  Each dot represents the performance of one trained predictor at a single regularization strength in terms of APL (x-axis, lower is better) and F1 score (y-axis, higher is better) on the heldout test set. Curves are formed by sweeping over a logarithmically-spaced range of strength values $\lambda$ with three runs each. Points closer to the top-left corner are the most simulable and performant minima.}
  \label{fig:uci}
\end{figure*}

\begin{figure*}[h!]
  \centering
  \begin{subfigure}[b]{\textwidth}
    \includegraphics[width=\textwidth]{uci_legend_final.pdf}
  \end{subfigure}
  \begin{subfigure}[b]{0.24\textwidth}
    \includegraphics[width=\textwidth]{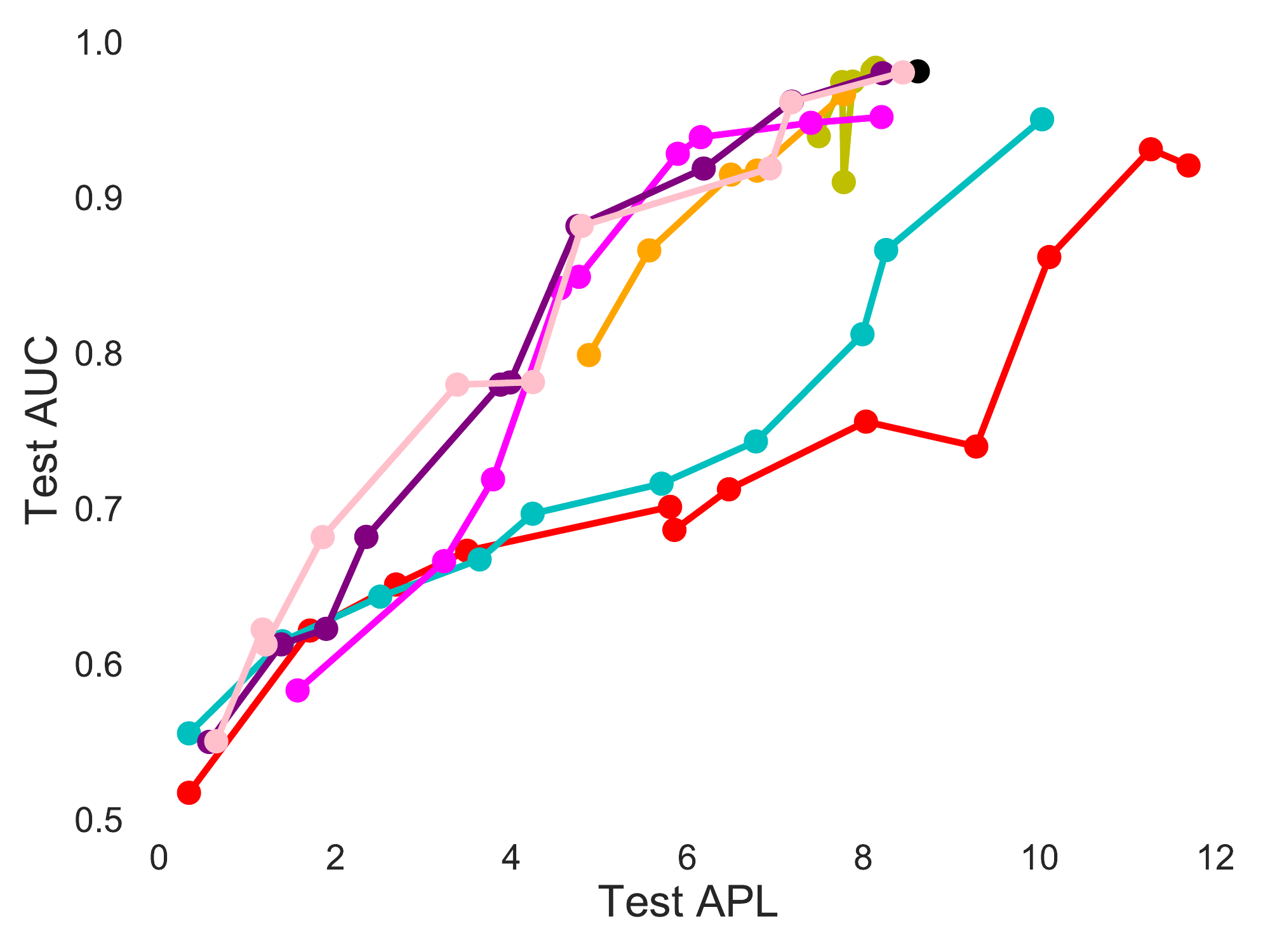}
    \caption{SOFA: Vasopressor}
  \end{subfigure}
  \begin{subfigure}[b]{0.24\textwidth}
    \includegraphics[width=\textwidth]{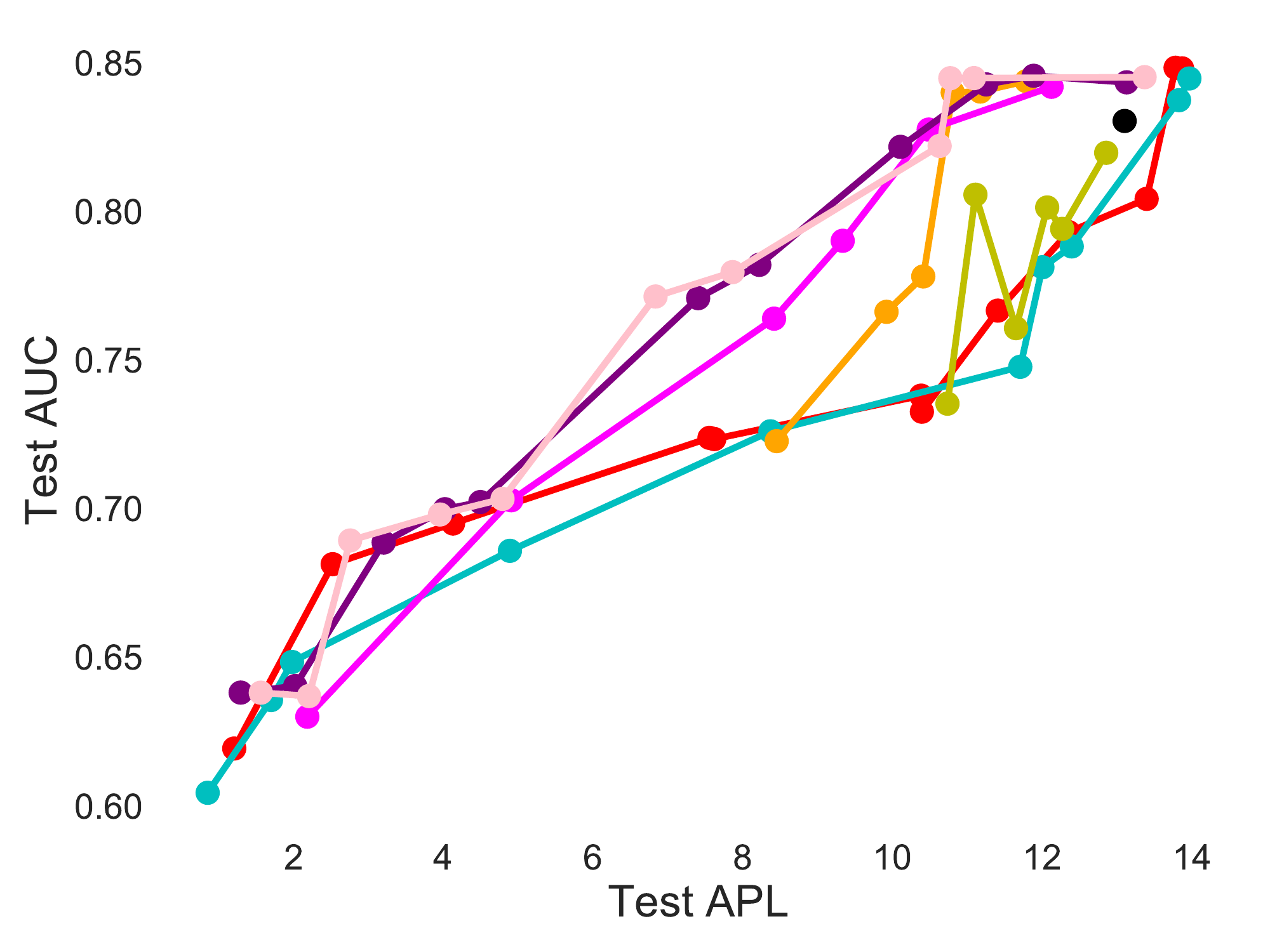}
    \caption{SOFA: Sedation}
  \end{subfigure}
  \begin{subfigure}[b]{0.24\textwidth}
    \includegraphics[width=\textwidth]{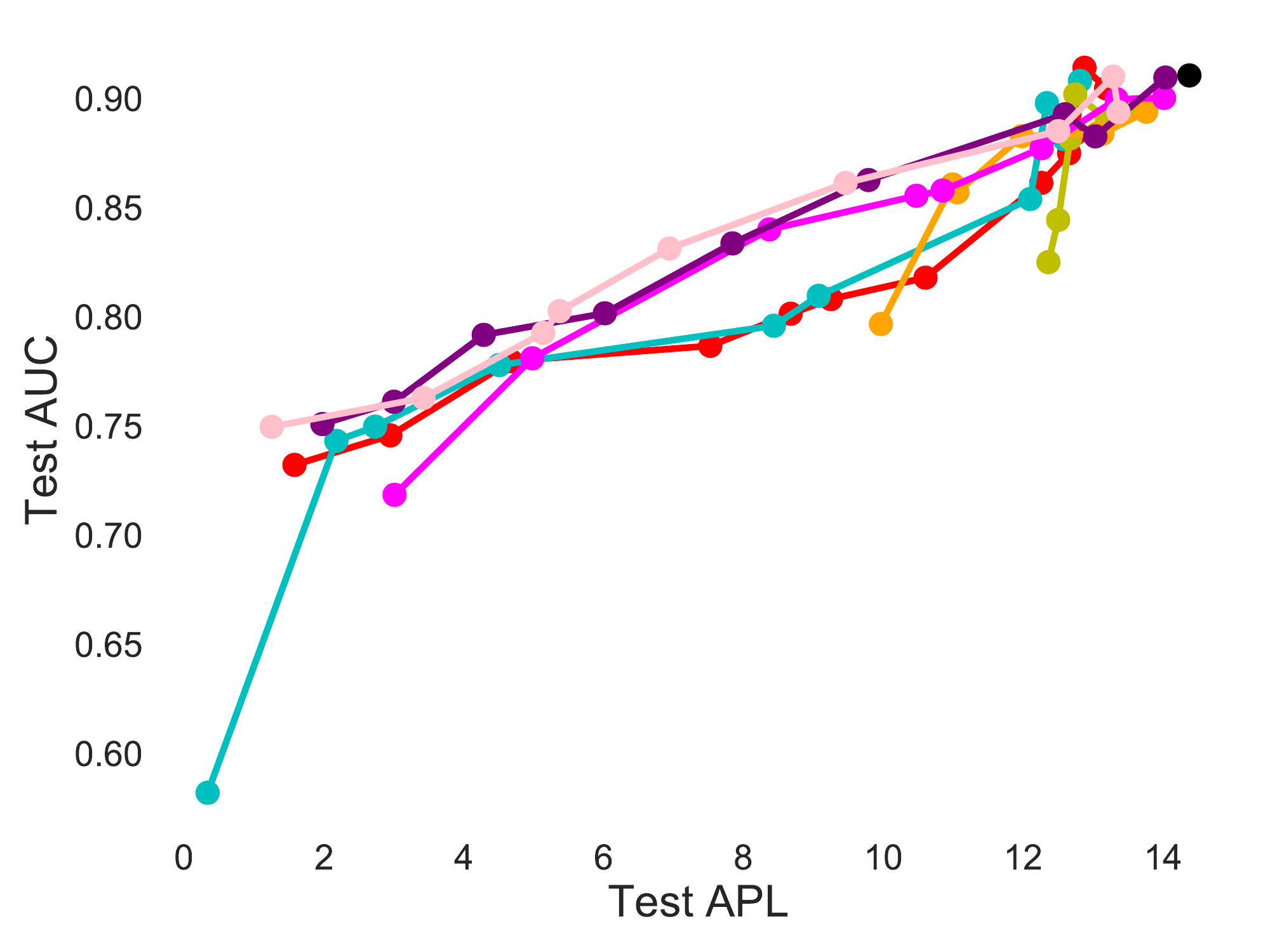}
    \caption{SOFA: Ventilation}
  \end{subfigure}
  \begin{subfigure}[b]{0.24\textwidth}
    \includegraphics[width=\textwidth]{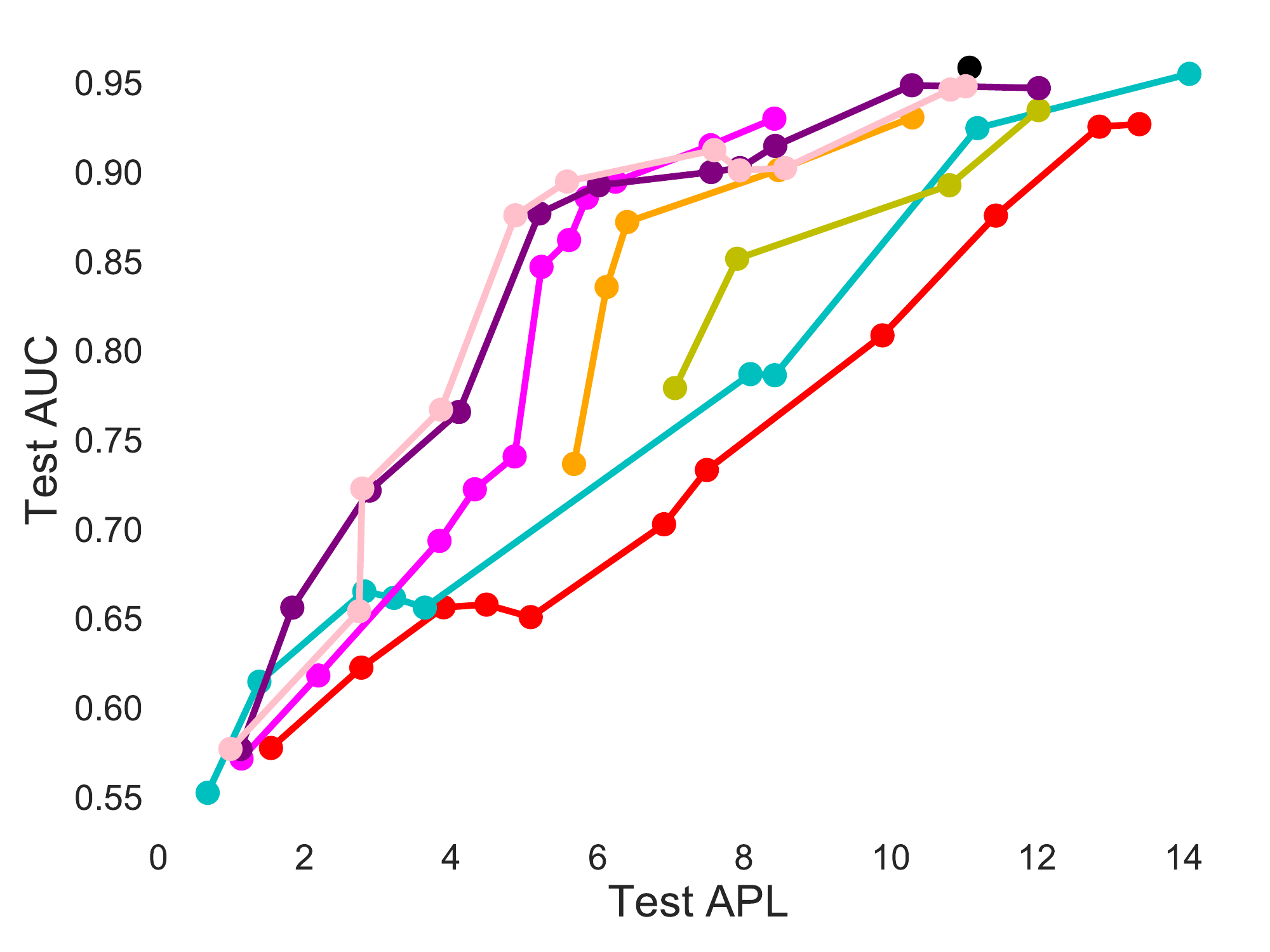}
    \caption{SOFA: Renal Therapy}
  \end{subfigure}
  \begin{subfigure}[b]{0.24\textwidth}
    \includegraphics[width=\textwidth]{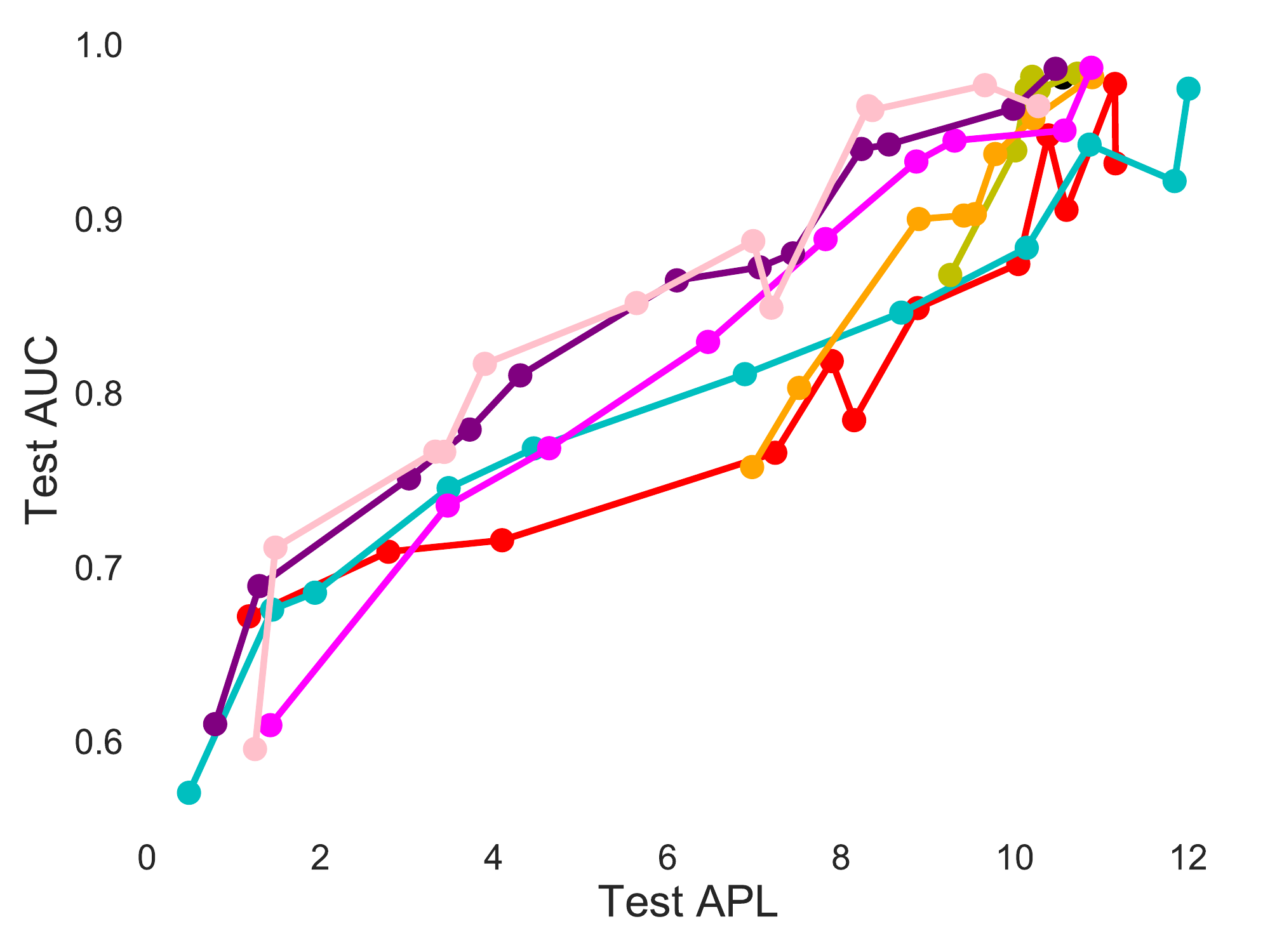}
    \caption{Careunit: Vasopressor}
  \end{subfigure}
  \begin{subfigure}[b]{0.24\textwidth}
    \includegraphics[width=\textwidth]{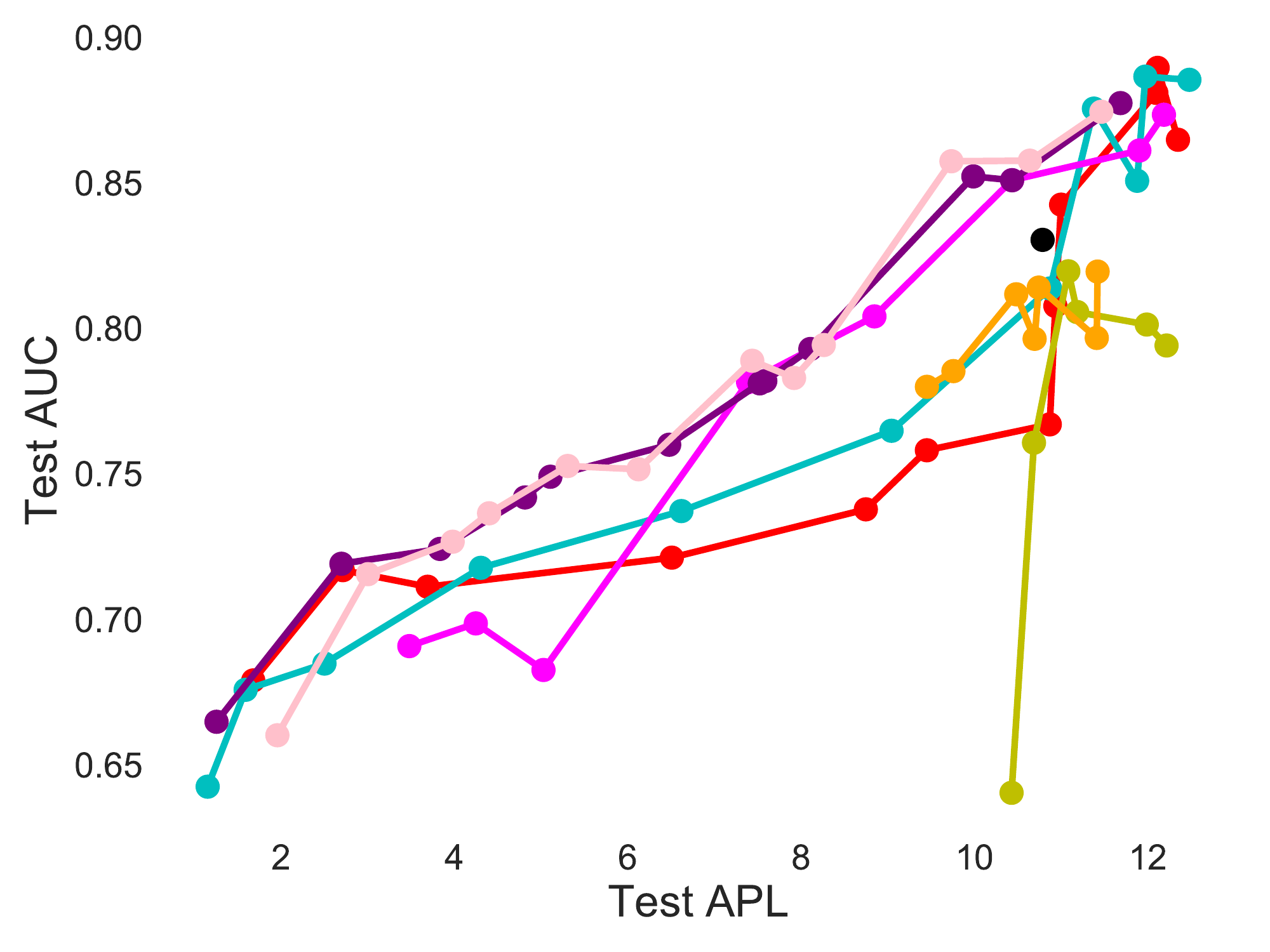}
    \caption{Careunit: Sedation}
  \end{subfigure}
  \begin{subfigure}[b]{0.24\textwidth}
    \includegraphics[width=\textwidth]{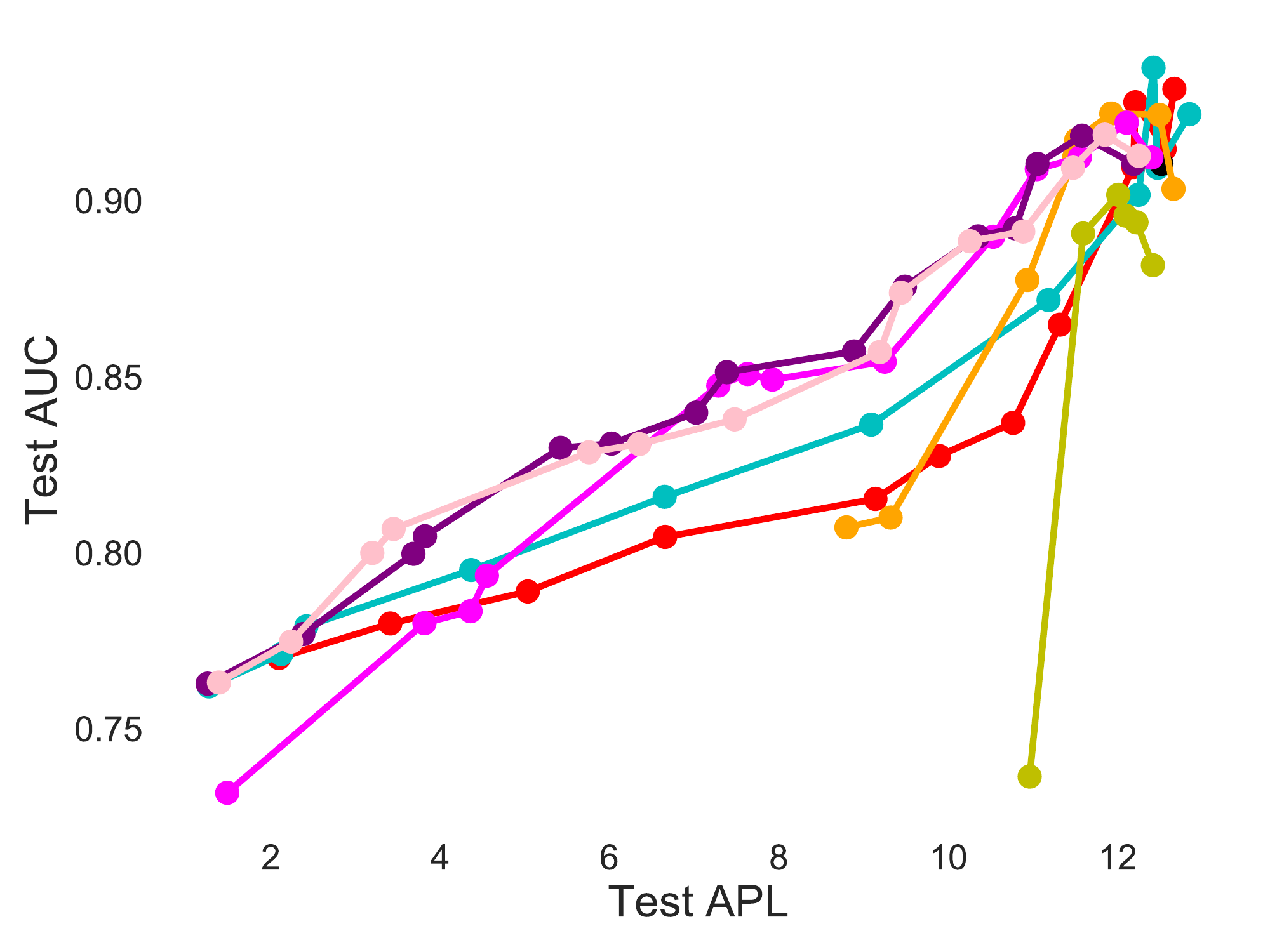}
    \caption{Careunit: Ventilation}
  \end{subfigure}
  \begin{subfigure}[b]{0.24\textwidth}
    \includegraphics[width=\textwidth]{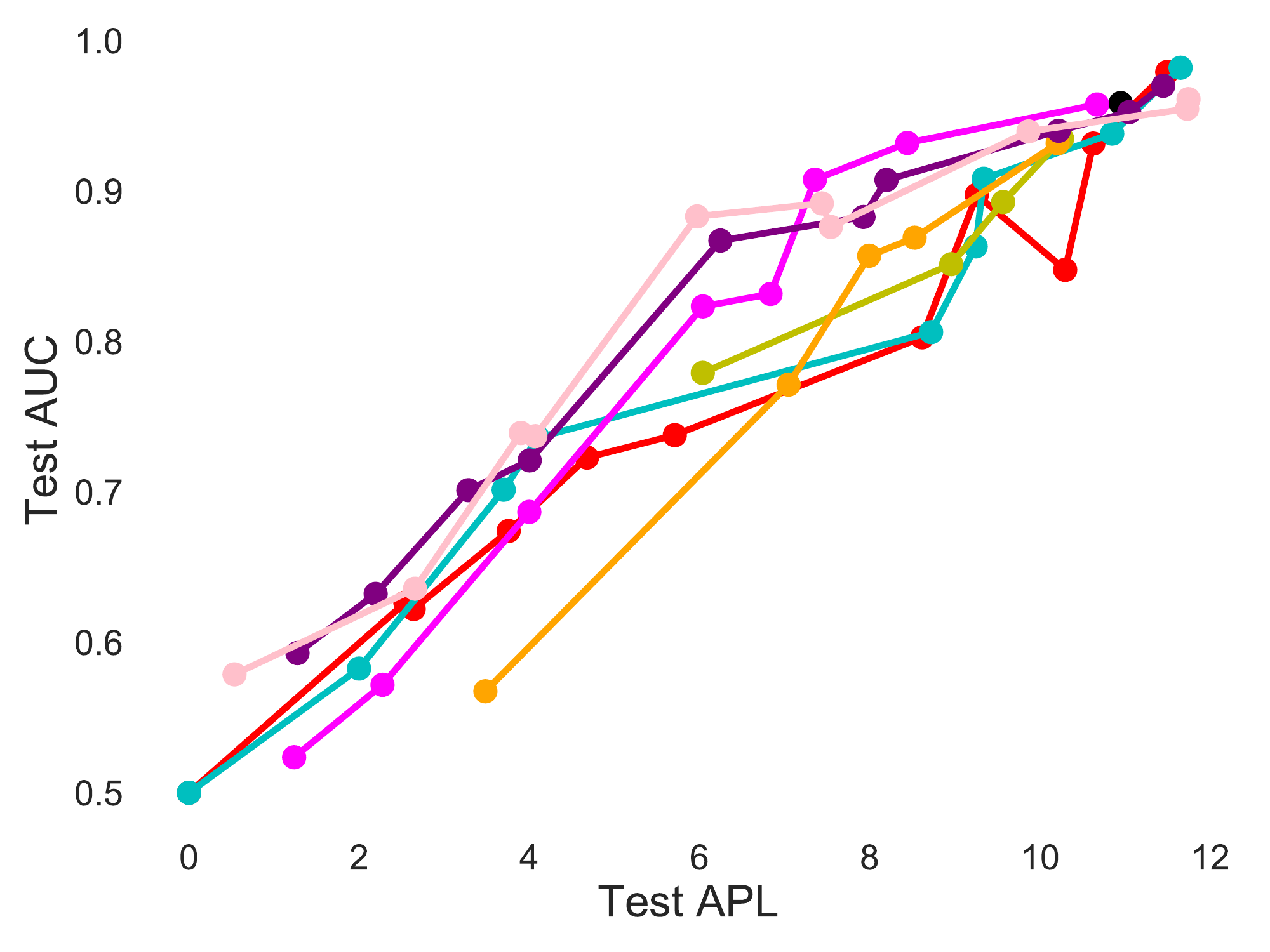}
    \caption{Careunit: Renal Therapy}
  \end{subfigure}
  \begin{subfigure}[b]{0.24\textwidth}
    \includegraphics[width=\textwidth]{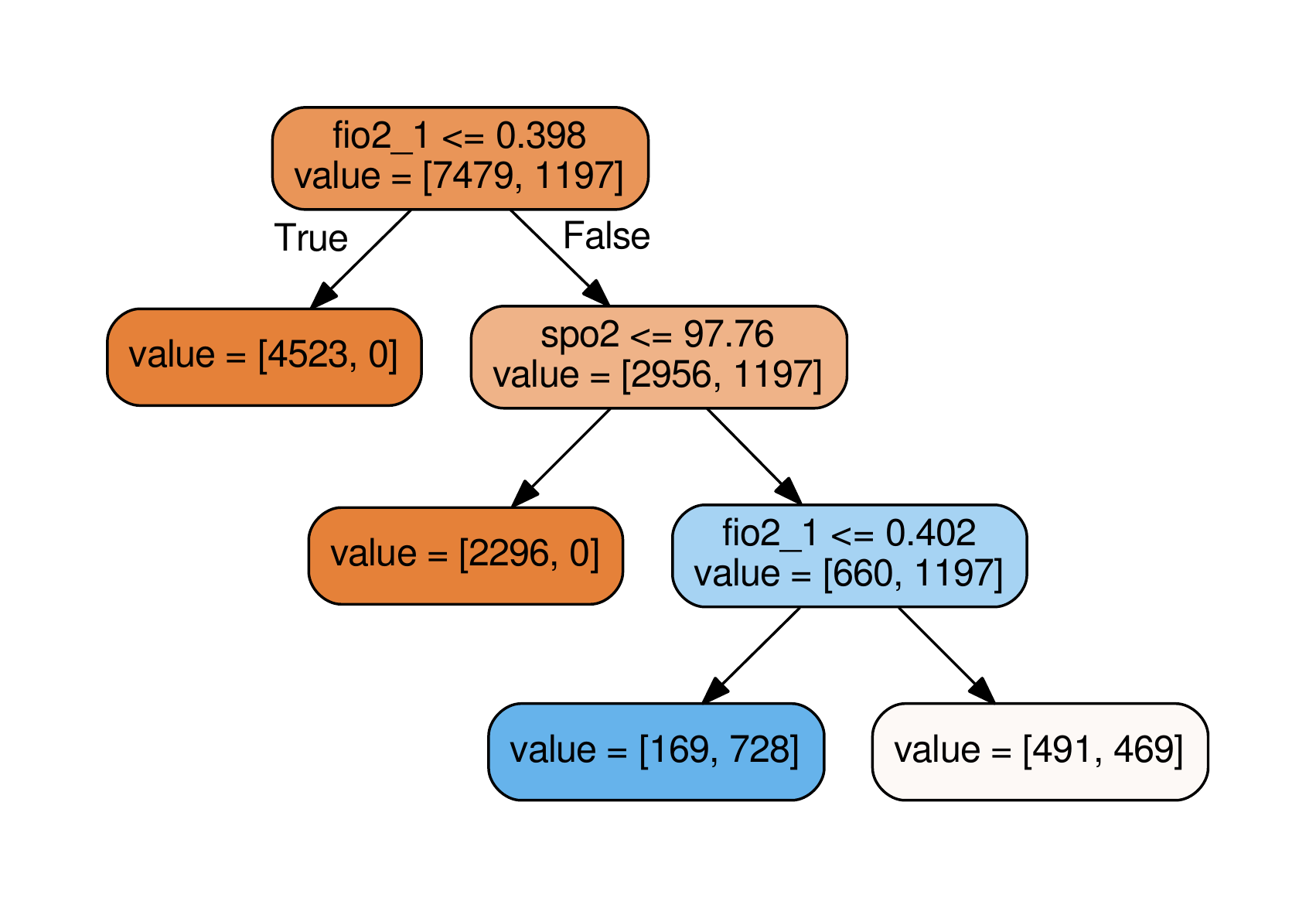}
    \caption{Low SOFA: Ventilation}
  \end{subfigure}
  \begin{subfigure}[b]{0.24\textwidth}
    \includegraphics[width=\textwidth]{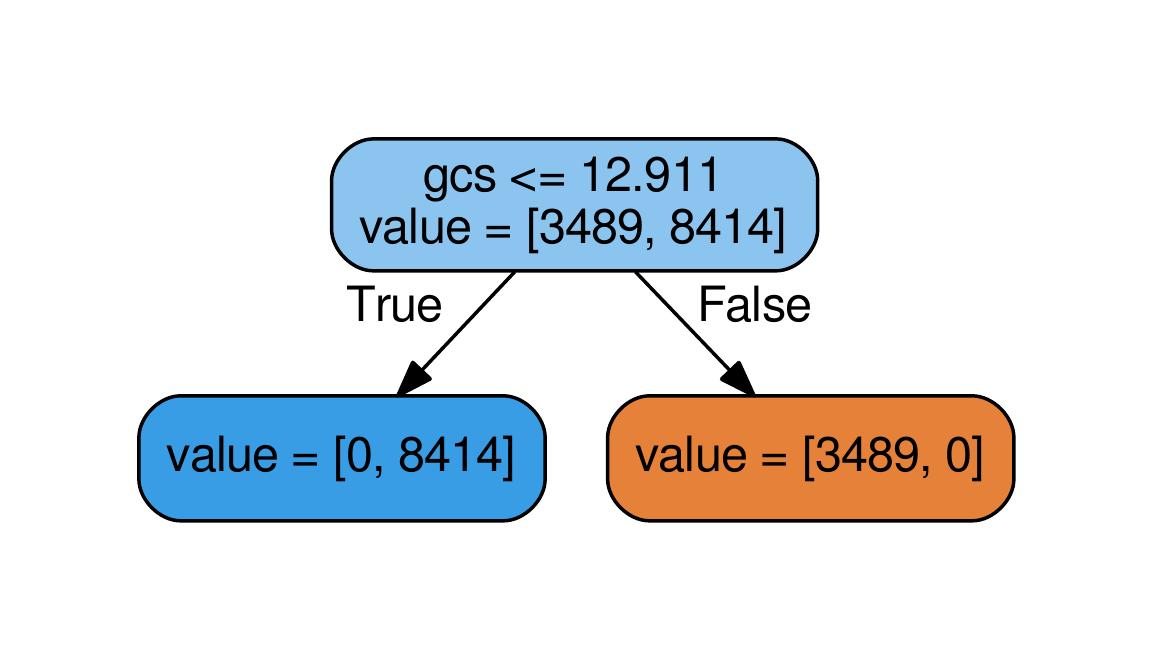}
    \caption{High SOFA: Ventilation}
  \end{subfigure}
  \begin{subfigure}[b]{0.24\textwidth}
    \includegraphics[width=\textwidth]{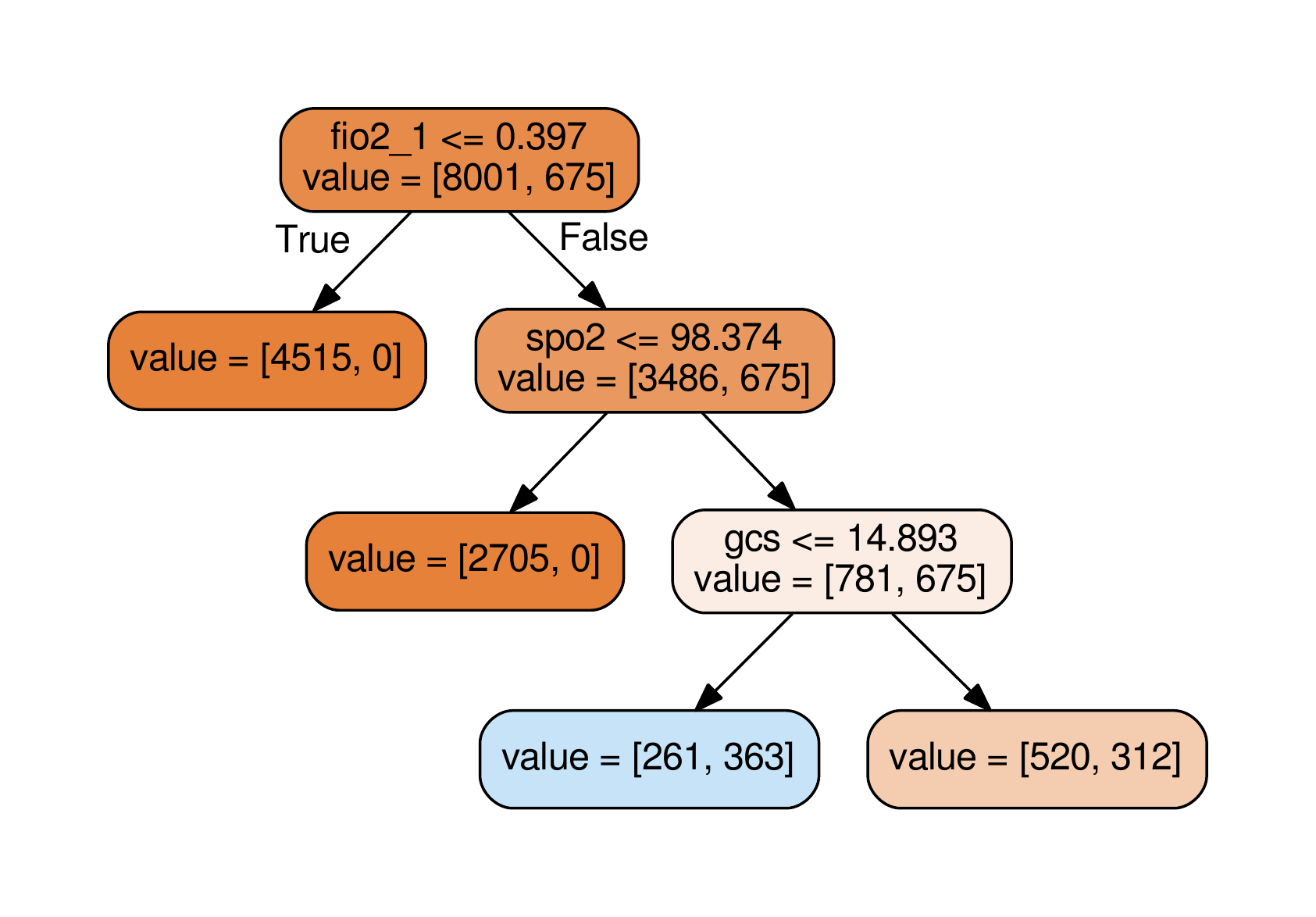}
    \caption{Low SOFA: Sedation}
  \end{subfigure}
  \begin{subfigure}[b]{0.24\textwidth}
    \includegraphics[width=\textwidth]{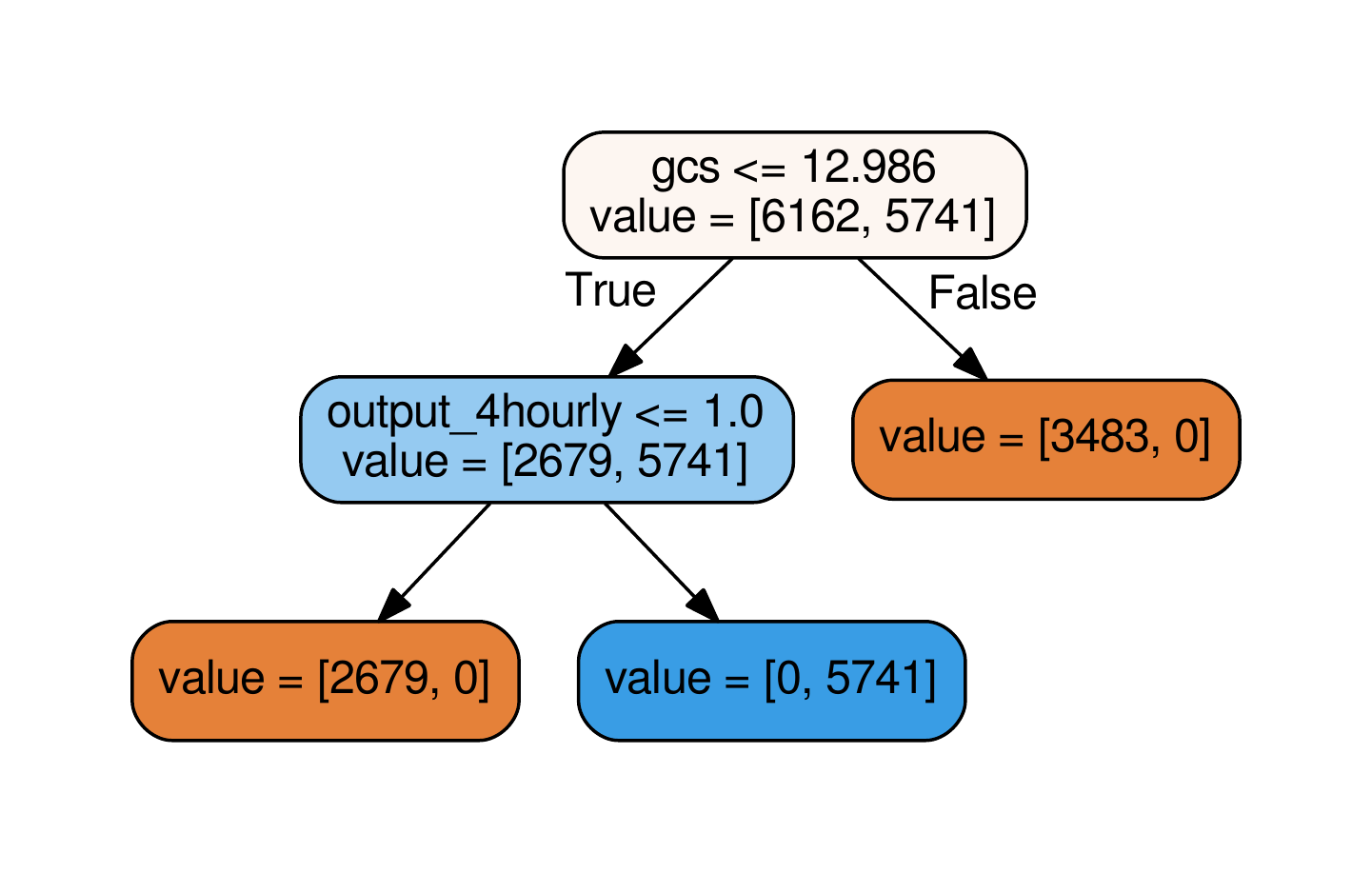}
    \caption{High SOFA: Sedation}
  \end{subfigure}
  \caption{Tradeoff curves for Critical Care. Each subfigure compares APL and test AUC (higher is better) for a different medication given in the intensive care unit. We compute APL using regions split by 3 SOFA scores (a-d) and 5 careunits (e-h). Finally, (i-l) show distilled trees that closely approximate a regional tree regularized model with low APL and high AUC.}
  \label{fig:sepsis}
\end{figure*}

\begin{figure*}[h!]
  \centering
  \begin{subfigure}[b]{\textwidth}
    \includegraphics[width=\textwidth]{uci_legend_final.pdf}
  \end{subfigure}
  \begin{subfigure}[b]{0.24\textwidth}
    \includegraphics[width=\textwidth]{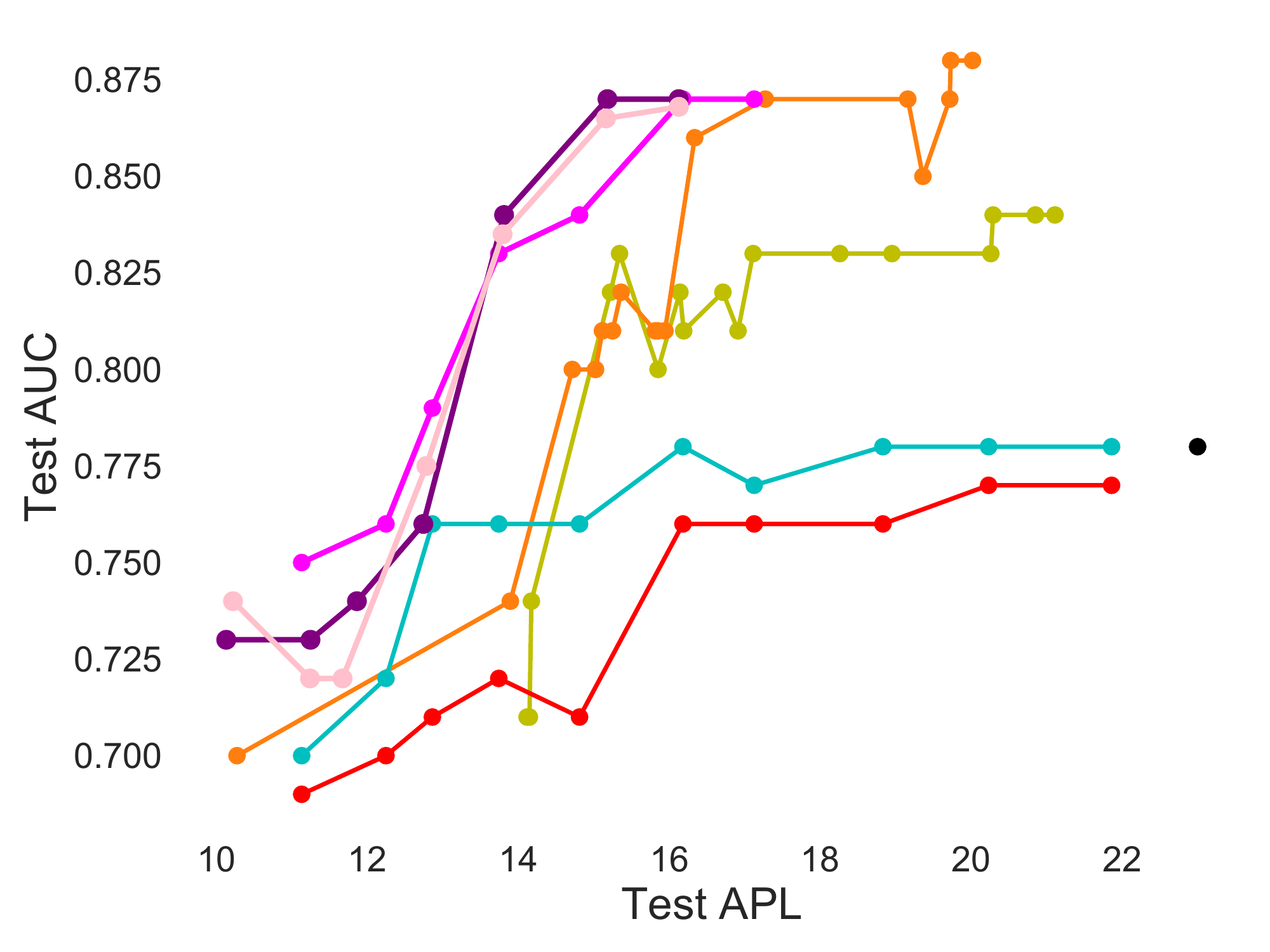}
    \caption{Immunity: Mortality}
     \label{mortalsubfig}
  \end{subfigure}
  \begin{subfigure}[b]{0.24\textwidth}
    \includegraphics[width=\textwidth]{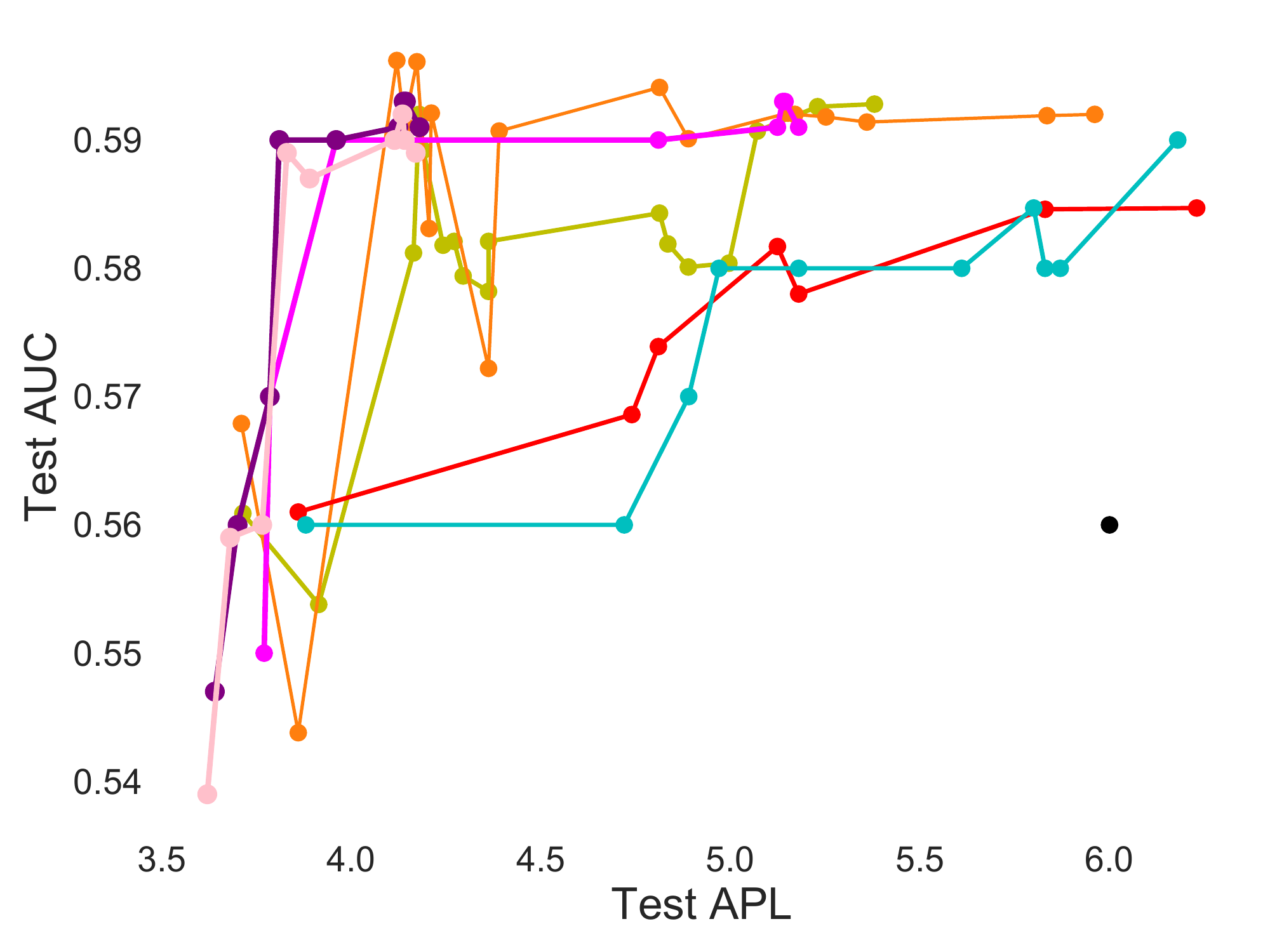}
    \caption{Immunity: AIDS Onset}
  \end{subfigure}
  \begin{subfigure}[b]{0.24\textwidth}
    \includegraphics[width=\textwidth]{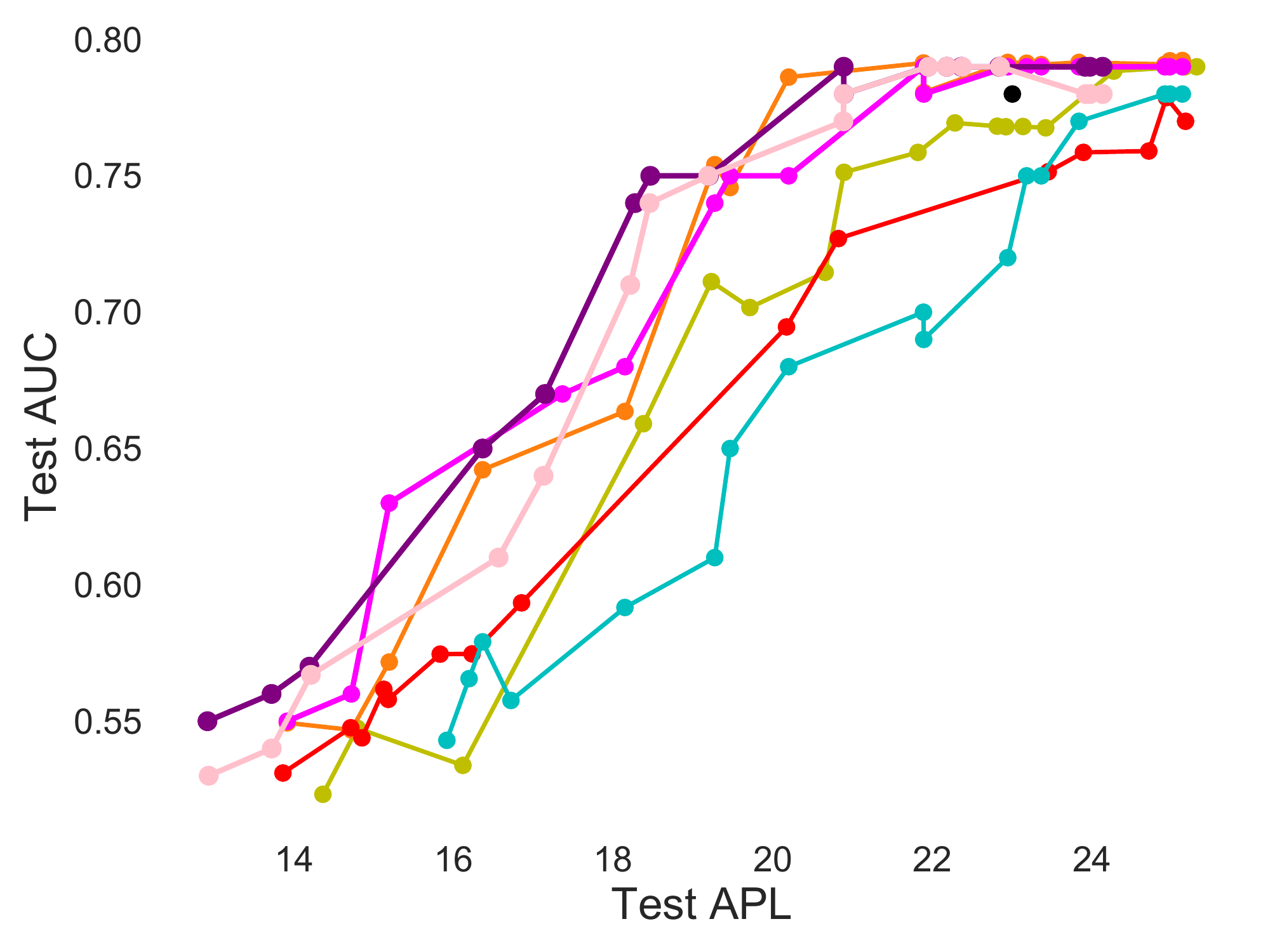}
    \caption{Immunity: Adherence}
  \end{subfigure}
  \begin{subfigure}[b]{0.24\textwidth}
    \includegraphics[width=\textwidth]{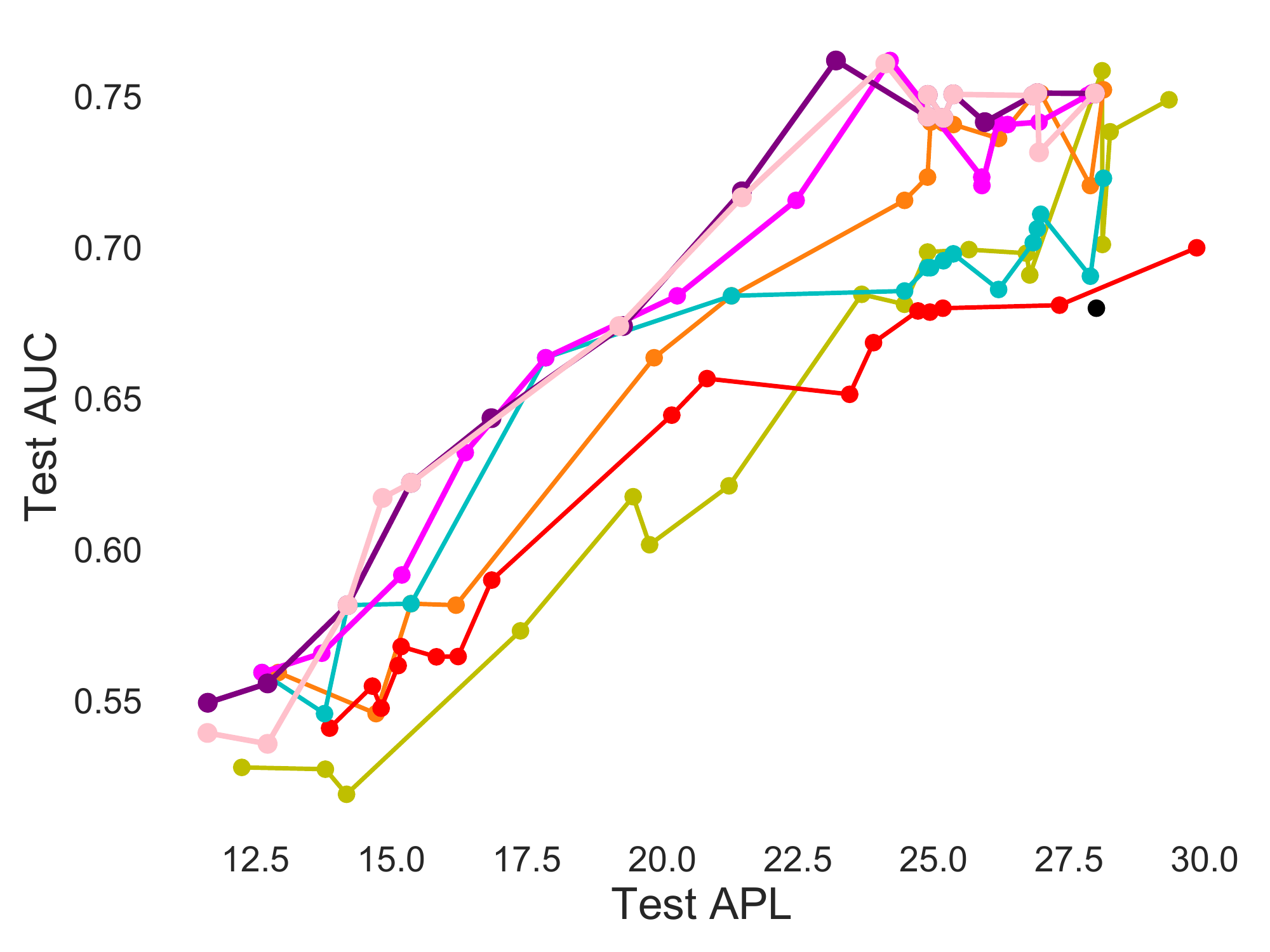}
    \caption{Immunity: Viral Suppr. }
    \end{subfigure}
     \begin{subfigure}[b]{0.32\textwidth}
    \includegraphics[width=\textwidth]{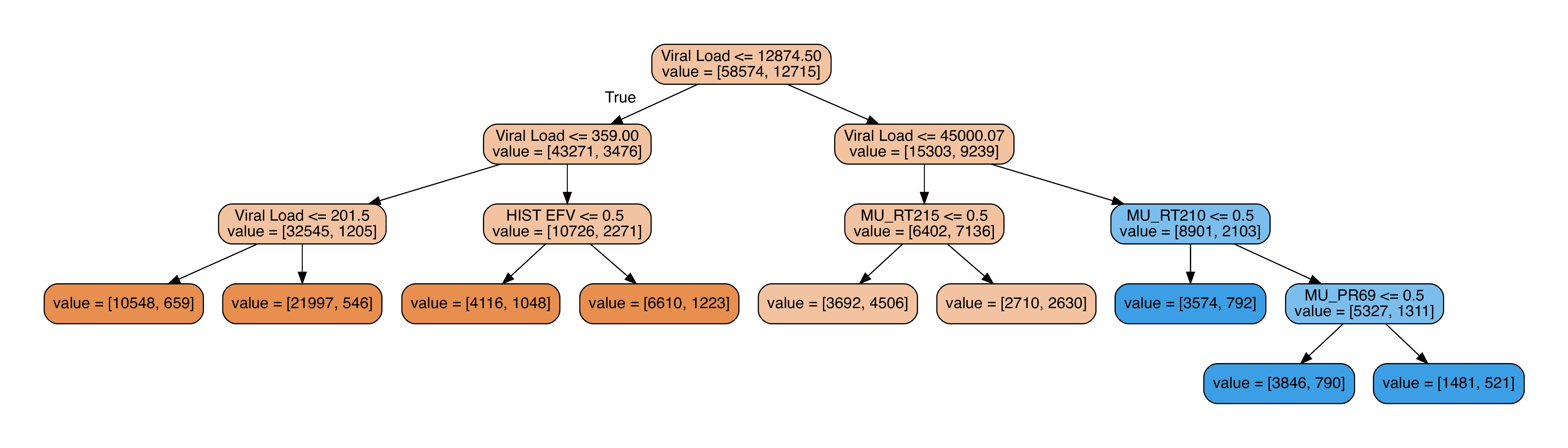}
    \caption{High Immunity: Mortality}
  \end{subfigure}
  \begin{subfigure}[b]{0.32\textwidth}
    \includegraphics[width=\textwidth]{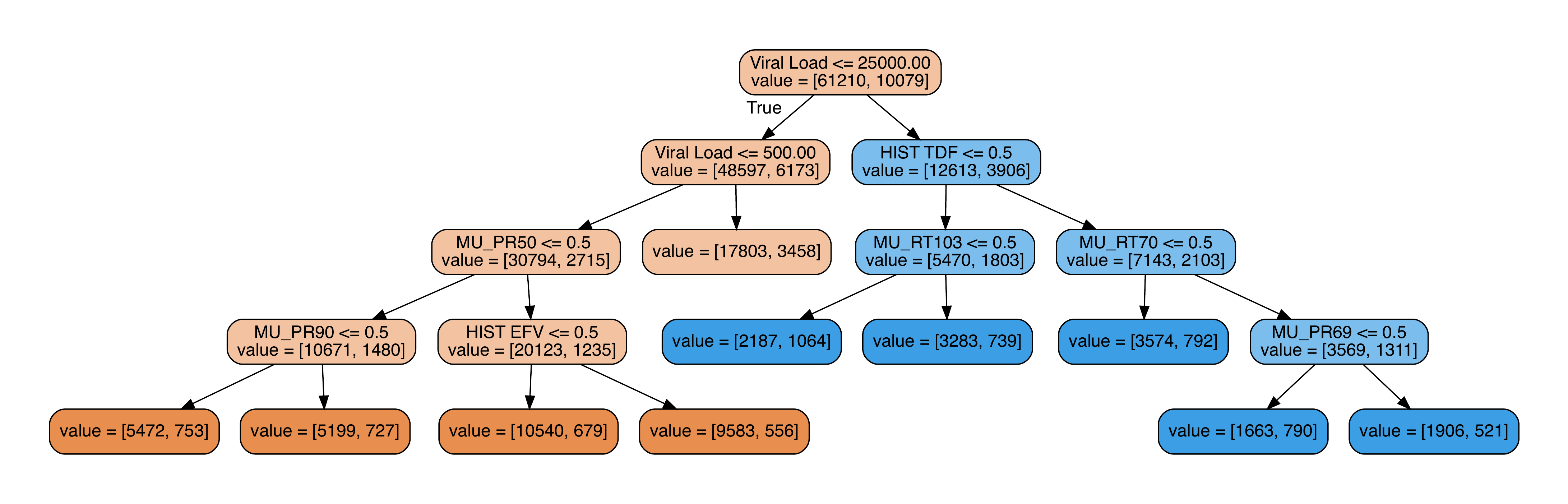}
    \caption{Mid Immunity: Mortality}
  \end{subfigure}
  \begin{subfigure}[b]{0.32\textwidth}
    \includegraphics[width=\textwidth]{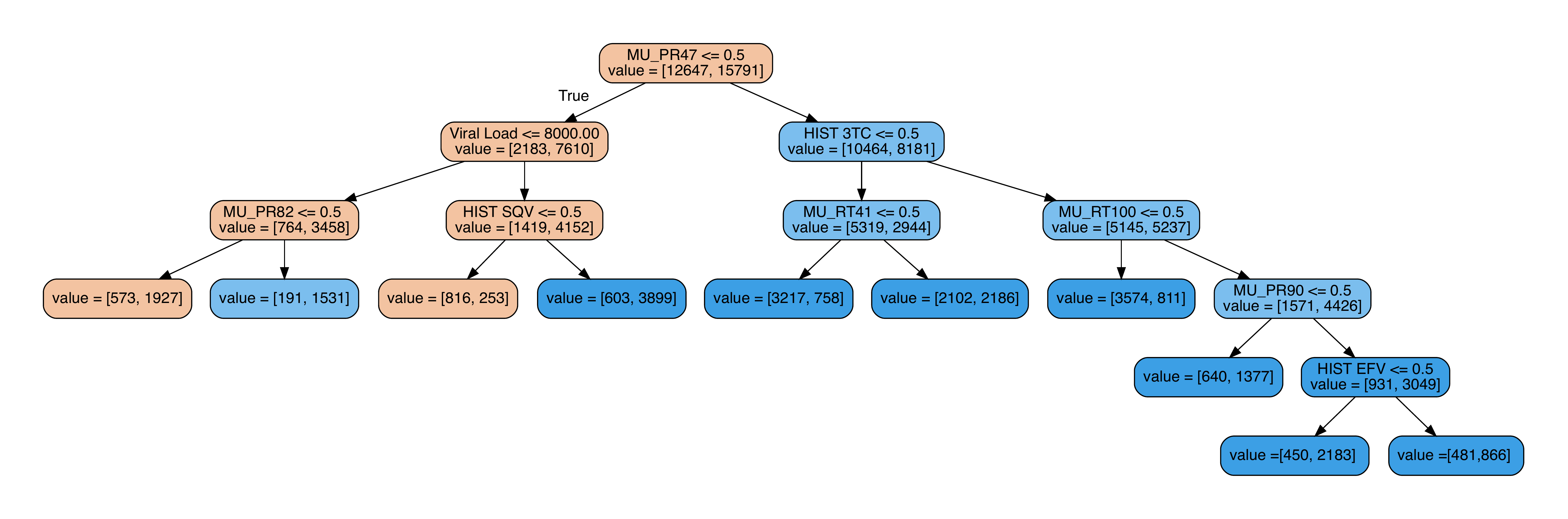}
    \caption{Low Immunity: Mortality}
  \end{subfigure}
  \caption{Tradeoff curves on the HIV dataset. We split regions by the level of immunosuppression (abbreviated to immunity) at baseline (e.g. $<$200 cells/mm$^3$). Subfigures (e-g) show distilled trees (confirmed to be simulable by an expert).}
  \label{fig:hiv}
\end{figure*}

\subsection{Demonstration on a Toy Example}
\label{sec:toy}

We first investigate a toy setting with a ground-truth function composed of five rectangles where each one is either shifted up or down (see Fig.~\ref{fig:toy}(a,b)). The training dataset is sparse (250 points) while the test dataset is much denser (5000 points). Noise is added to the training labels to encourage overfitting. This is intended to model real-world settings where regional structure is only partially observable from an empirical dataset. It is exactly in these contexts that regularization can help. See Appendix for more details.
\begin{table}[b]
\centering
\small
\caption{Classification accuracy on a toy example. The reported test APL is averaged over APLs in five regions.}
\begin{tabular}{c c c c}
\toprule
Model & Test Acc. & Test APL \\
\midrule
Unregularized & $0.8296$ & $17.9490$ \\
L$_2$ ($\lambda=0.001$) & $0.8550$ & $16.1130$ \\
Global Tree ($\lambda=1$) & $0.8454$ & $\mathbf{6.3398}$ \\
L$_1$ Regional Tree ($\lambda=0.1$) & $0.9168$ & $10.1223$ \\
L$_0$ Regional Tree ($\lambda=0.1$) & $0.9287$ & $8.1020$ \\
L$_{\text{SP}}$ Regional Tree ($\lambda=0.1$) & $\mathbf{0.9308}$ & $8.1962$ \\
\bottomrule
\end{tabular}
\label{table:toy}
\end{table}
Fig.~\ref{fig:toy} show the learned boundary with various regularizers. As global regularization is restricted to penalizing all data points evenly, increasing the strength causes the target neural model to collapse from a complex boundary to a single axis-aligned boundary (e). Similarly, if we increase the strength of L$_2$ regularization even slightly from (d), the model collapses to the trivial solution. Only regional tree regularization (f,g) is able to model the up-and-down curvature of the true function. With high strength, L$_\text{SP}$ regional tree regularization produces a more axis-aligned boundary than L$_1$, primarily because we can regularize complex regions more harshly without collapsing simpler regions.

Table~\ref{table:toy} compares classification accuracy: regional tree regularization achieves the lowest error while remaining simulable. While global tree regulariziation finds the minima with lowest APL, this comes at the cost of accuracy. With any regularizer, we could have chosen a high enough penalty such that the test APL would be 0, but the resulting accuracy would approach chance. The results in Table~\ref{table:toy} show that regional regularizers find a good compromise between accuracy and complexity.

\subsection{UC Irvine Repository}
\label{sec:experiments}
We now apply regional tree regularization to four  datasets from the UC Irvine repository \citep{Dua:2017}. We will refer to these as \textit{Bank}, \textit{Gamma}, \textit{Adult}, and \textit{Wine}. See Appendix for a description of each. We choose a generic method for defining regions to showcase the wide applicability of regional regularization: we fit a $k$-means clustering model with $k=5$ to each dataset.

Fig.~\ref{fig:uci} compares F1 scores and APL for each dataset. First, we can see that an unregularized model (black) does poorly due to overfitting. Second, we find that (as expected) a penalty on the L$_2$ norm is not a good regularizer for simulability, as it is unable to find many minima in the low APL region (see Gamma, Adult, and Wine under roughly 5 APL). Any increase in strength quickly causes the target neural model to degenerate to predicting a single label (an F1 score of 0). Interestly, we see similar behavior with global tree regularization, suggesting that finding low complexity minima is challenging under strong global constraints. As an additional benchmark, we tried global tree regularization where the region index (1 to $R$) for each data point is appended to the feature vector. We did not find this change to improve performance nor simulability. Third, regional tree regularization achieves the highest test accuracy in all datasets. With low APL, regional explanations surpasses global explanations in performance. For example, in Bank, Gamma, Adult, and Wine, we can see this at 3-6, 4-7, 5-8, 3-4 APL respectively.
Under very high strengths, regional tree regularization converges in performance with regional decision trees, which is sensible as the neural network focuses on distillation. Finally,  consistent with toy examples, L$_0$/L$_\text{SP}$ regional tree regularization finds more performant minima with low to mid APL than L$_1$. We believe this to largely be due to ``evenly" regularizing regions.

\section{Healthcare Applications}
We turn to two real-world use cases: predicting interventions in critical care and predicting HIV medication usage. The critical care task, performed with the MIMIC dataset \citep{johnson2016mimic}, involves taking a patient's current statistics and predicting whether they are undergoing 4 different kinds of therapies (vasopressor, sedation, ventilation, renal therapy).  Regions are constructed based on how our intensivist collaborators described dividing the acuity (SOFA) and treatment unit (surgical vs. medical) of the patient.  The HIV task, performed with the EUResist dataset \citep{zazzi2011prediction}, also takes in patient statistics and now predicts 15 outcomes having to do with drug response. In consultation with clinical experts, the regions reflect different levels of immunosuppression. The details of the data are included in the Appendix; below we highlight the main results.

\paragraph{L$_\text{SP}$ regional tree regularization finds a wealth of simulable solutions with high accuracy.}
A performant and simulable model has high F1/AUC scores near a low APL. Across experiments, we see regional tree regularization is most adept at finding such minima: in each subfigure of Fig.~\ref{fig:sepsis} and Fig.~\ref{fig:hiv}, we can point to a span of APL at which the pink curves are much higher than all others. Further, the sparsity induced by L$_0$ norm helped find even more desirable minima than with any ``dense norm" (L$_1$, L$_2$, etc). As evidence, in low APL regions, the dotted pink lines contain points above all others. In constrast, global regularizers struggled, likely due to strong constraints that made optimization difficult. We can see evidence for this in Critical Care: in Fig.~\ref{fig:sepsis} (a,c,e), the minima from global constraints stay very close to unregularized minima. Even worse, in (f, g), global regularizers settle for bad optima, reaching low accuracy with high APL. We observe similar findings for HIV in Fig~\ref{fig:hiv} where global explanations lead to poor accuracies.

\paragraph{L$_\text{SP}$ and L$_0$ converge to similar minima but the former is much faster.}
In every experiment, the AUC/F1 and APL of the minima found by L$_0$ and L$_{\text{SP}}$ regularized deep networks are close to identical. While this suggests the two regularizers are comparable, using $\texttt{sparsema}$ instead of $\texttt{max}$ speeds up training ten-fold (in terms of epochs).


\paragraph{Regional tree regularization distills decision trees for each region.}
For Critical Care, Fig.~\ref{fig:sepsis}(i,j) show region-specific decision trees for the need-for-ventilator task selected from a low APL and high AUC minima of a regional tree regularized model. The structure of the trees are different, indicating the decision logic changes substantially for low- and high-risk regions. Moreover, while Fig.~\ref{fig:sepsis}(i) mostly predicts 0 (orange), Fig.~\ref{fig:sepsis}(j) mostly predicts 1 (blue), which agrees with common intuition that SOFA scores are correlated with mortality.
We see similar patterns from the distilled trees for HIV: Fig~\ref{fig:hiv}(e-g). In particular, we observe that lower levels of immunity at baseline are associated with higher viral loads and risk of mortality. If we were to use a single tree, we would either lose granularity or simulability.

\paragraph{Distilled decision trees are clinically useful.} We asked for feedback from specialist clinicians to assess the trees produced by regional tree regularization. For Critical Care, an  intensivist noted that the explanations allowed him to connect the model to his cognitive categories of patients. For example, he verified that for predicting ventilation, GCS (mental status) was indeed a key factor.
Moreover, he was  able to make useful requests: he asked if the effect of oxygen could have been a higher branch in the tree to better understand its effects on ventilation choices, and, noticing the similarities between the sedation and ventilation trees, suggested defining new regions by both SOFA and ventilation status. For HIV, a second clinician confirmed our observations about relationships between viral loads and mortality. He also noted that when patients have lower baseline immunity, the trees for mortality contain several more drugs. This is consistent with medical knowledge, since patients with lower immunity require more aggressive therapies to combat drug resistance. We also showed the clinician two possible trees for a high-risk region: the first was from regional tree regularisation; the second was from a decision tree trained using data from only this region. The clinician preferred the first tree as the decision splits captured more genetic information about the virus that could be used to reason about resistance patterns to antiretroviral therapy.



\paragraph{Regional tree regularizers make faithful predictions.}
Table~\ref{table:fidelity} shows the fidelity of a deep model to its distilled decision tree. A score of 1.0 indicates that both models learned the same decision function, which is actually undesirable. The ``perfect" model would have high but not perfect fidelity, disagreeing with the decision tree a small portion of the time. With an average fidelity of 89\%, the distilled tree is trustworthy as an explanatory tool in most cases, but can take advantage of deep nonlinearity with difficult examples.
\begin{table}[h!]
\centering
\small
\caption{Fidelity shows the percentage of examples on which the prediction made by a tree agrees with the deep model.}
\begin{tabular}{c c c c c c}
\toprule
Bank & Gamma & Adult & Wine & Critical Care & HIV \\
\midrule
0.892 & 0.881 & 0.910 & 0.876 & 0.900 & 0.897 \\
\bottomrule
\end{tabular}
\label{table:fidelity}
\end{table}
Fidelity is also controllable by the regularization strength. With a high penalty, fidelity willl near 1 at the price of accuracy. It is up to the user and domain to decide what fidelity is best.

\paragraph{Per-Epoch Computational Cost Comparison} Averaging over 100 trials on Critical Care, an L$_2$ model takes $2.39 \pm 0.2$ sec. per epoch. Global tree models take $5.90 \pm 0.4$ sec. to get 1\,000 training samples for the surrogate network using data augmentation and compute APL for $\mathcal{D}^\theta$, and $21.42\pm0.6$ sec. to train the surrogate model for 100 epochs. Regional tree models take $6.603\pm0.271$ sec. and $39.878\pm0.512$ sec. respectively for 5 surrogates. The increase in base cost is due to the extra forward pass through $R$ surrogate models to predict APL. The surrogate cost(s) are customizable depending on the size of $\mathcal{D}^\theta$, the number of training epochs, and the frequency of re-training. If $R$ is large, we need not re-train each surrogate; we can randomly sample regions. Surrogate training can be parallelized.

\section{Discussion and Conclusion}
Interpretability is a bottleneck preventing widespread acceptance of deep learning.  In this work, we introduced regional tree regularization, which enforces that a neural network is simple across \emph{all} expert-defined regions---something that previous regularizers could not do. While we used relatively simple ways to elicit regions from experts, future work could iterate between using our innovations in how to optimize networks for simple regional explanations given regions and using interactive methods to elicit regions from experts.

\section{Acknowledgements}
MW is supported by NSF GRFP. SP is supported by the Swiss National Science Foundation projects 51MRP0$\_$158328 and P2BSP2$\_$184359. MW and FDV acknowledge support from a Sloan Fellowship. The authors thank the EuResist Network for providing HIV data and Matthieu Komorowski for sepsis data. Computations were supported by the FAS Research Computing Group at Harvard and sciCORE (http://scicore.unibas.ch/) scientific computing core facility at University of Basel.

\section{Appendix}
Supplement is available at
\url{https://arxiv.org/abs/1908.04494}. PyTorch implementation is available at \url{https://github.com/mhw32/regional-tree-regularizer-public}.

\bibliographystyle{aaai}
\bibliography{AAAI-WuM.2899}

\end{document}